\documentclass[11pt]{article}

\usepackage[utf8]{inputenc} 
\usepackage[T1]{fontenc}    
\usepackage{hyperref}       
\usepackage{url}            
\usepackage{booktabs}       
\usepackage{amsfonts}       
\usepackage{nicefrac}       
\usepackage{microtype}      
\usepackage{xcolor}         
\usepackage{adjustbox}      
\usepackage{makecell}

\usepackage[final]{acl}
\usepackage{times}
\usepackage{latexsym}
\usepackage{inconsolata}

\usepackage{listings}
\usepackage{amsmath}
\usepackage{graphicx}
\usepackage{mdframed}
\usepackage{framed}
\usepackage{wrapfig}
\usepackage{subcaption}

\usepackage{etoolbox}
\usepackage{enumitem}

\newcommand{\task}{\texttt{{T}}}
\newcommand{\code}{\texttt{{C}}}
\newcommand{\goal}{\texttt{G}}
\newcommand{\cons}{{\texttt{L}}}
\newcommand{\grid}{\texttt{W}}

\newcommand{\codecons}{\textcolor{blue}{\texttt{A}}}
\newcommand{\gridcons}{\textcolor{cyan}{\texttt{B}}}
\newcommand{\spatial}{\textcolor{red}{\texttt{C}}}

\newcommand{\fd}{\texttt{move\_forward()}}
\newcommand{\bk}{\texttt{move\_back()}}
\newcommand{\lt}{\texttt{turn\_left()}}
\newcommand{\rt}{\texttt{turn\_right()}}
\newcommand{\setpc}{\texttt{setpc}}
\newcommand{\Repeat}{\texttt{for}}
\newcommand{\atmost}{\texttt{AtMost}}
\newcommand{\startby}{\texttt{StartBy}}
\newcommand{\exactly}{\texttt{Exactly}}
\newcommand{\none}{\texttt{None}}
\newcommand{\hybrid}{\texttt{Hybrid}}

\newcommand{\find}{\texttt{Find}}

\newcommand{\counting}{\texttt{Math}}

\newcommand{\draw}{\texttt{Draw}}
\newcommand{\logic}{\texttt{Logic}}

\newcommand{\ProgBasic}{\text{Basic Actions}}
\newcommand{\ProgBasicRepeat}{\text{Loops}}
\newcommand{\ProgBasicSetpc}{\text{Variables}}
\newcommand{\ProgAll}{\text{Loops and Variables}}
\newcommand{\Hybrid}{\texttt{Hybrid}}

\newcommand{\diff}{\texttt{D}}
\newcommand{\easy}{\texttt{Easy}}
\newcommand{\medium}{\texttt{Medium}}
\newcommand{\hard}{\texttt{Hard}}

\newcommand{\platformAll}{XLogoOnline}
\newcommand{\platformMini}{XLogoOnline-Mini}
\newcommand{\benchmark}{\textsc{XLogoMiniProg}}

\newcommand{\benchmarkreal}{\textsc{Basic}}
\newcommand{\benchmarksim}{\textsc{Sim}}
\newcommand{\benchmarksimtrain}{\textsc{SimTrain}}
\newcommand{\benchmarksimval}{\textsc{SimVal}}
\newcommand{\benchmarksimeval}{\textsc{SimEval}}

\newcommand{\sft}{\text{Uni}} 
\newcommand{\emulator}{\text{Emu}}

\definecolor{promptinputcolor}{rgb}{0,0,2.55}
\newcommand{\promptheader}[1]{{\large{\textcolor{blue}{\textbf{#1}}}}}
\newcommand{\promptinput}[1]{{\textcolor{promptinputcolor}{{#1}}}}

\title{Program Synthesis Benchmark for Visual Programming \\ in XLogoOnline Environment}

\author{Chao Wen \\
MPI-SWS \\
\texttt{chaowen@mpi-sws.org} \\
\And
Jacqueline Staub \\
University of Trier \\
\texttt{staub@uni-trier.de} \\
\And
Adish Singla \\
MPI-SWS \\
\texttt{adishs@mpi-sws.org}
}

\begin{document}

\maketitle

\newtoggle{MainSuppContent}
\settoggle{MainSuppContent}{true}
  
\newtoggle{MainContentOnly}
\settoggle{MainContentOnly}{false}

\newtoggle{SuppContentOnly}
\settoggle{SuppContentOnly}{false}


\iftoggle{MainSuppContent}{

\begin{abstract}
Large language and multimodal models have shown remarkable success on various benchmarks focused on specific skills such as general-purpose programming, math word problem-solving, and visual question answering. However, it is unclear how well these models perform on tasks that require a combination of these skills. In this paper, we curate a novel program synthesis benchmark based on the real-world tasks in the XLogoOnline visual programming environment. Each task requires a combination of different skills such as spatial planning, basic programming, and logical reasoning. Our evaluation shows that current state-of-the-art models like GPT-4V and Llama3-70B struggle to solve these tasks, achieving only $20\%$ and $2.35\%$ success rates, respectively. Next, we develop a fine-tuning pipeline to boost the performance of models by leveraging a large-scale synthetic training dataset with over $80,000$ tasks. Moreover, we showcase how emulator-driven feedback can be used to design a curriculum over training data distribution, through which a fine-tuned Llama3-8B drastically outperforms GPT-4V and Llama3-70B models. Finally, we provide an in-depth failure analysis to understand the limitations of different models. We will publicly release the benchmark for future research on program synthesis in visual programming. 
\end{abstract}



\section{Introduction} \label{sec.intro}

\begin{figure}[ht!]
    \includegraphics[width=0.9\linewidth]{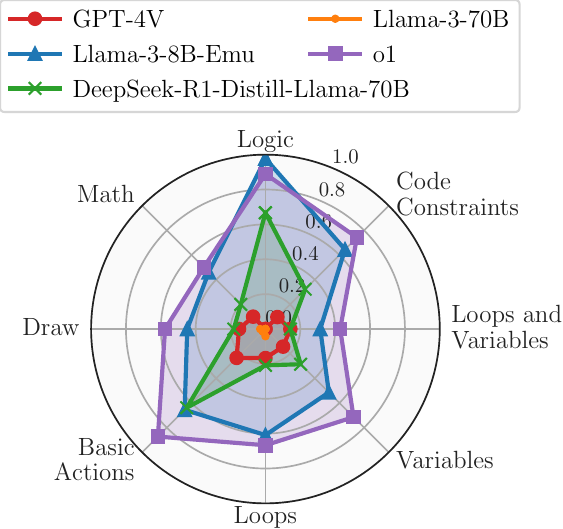}
    \caption{Success rates of different models across different skills in real-world tasks in the \benchmark{} benchmark.}
    \label{fig.intro.radar_real_skills}
\end{figure}

\looseness-1
In recent years, large models have shown remarkable performance in various domains, such as general-purpose programming and visual question answering~\citep{DBLP:journals/corr/abs-2303-12712}. For instance, in programming, numerous tools and models use large language models (LLMs) for code generation~\citep{DBLP:journals/corr/abs-2107-03374,githubcopilot} and programming feedback generation~\citep{DBLP:conf/lak/PhungPS0CGSS24,DBLP:conf/edm/PhungCGKMSS23,DBLP:conf/icer/PhungPCGKMSS22}, revolutionizing how programmers write code and how teachers instruct programming~\citep{DBLP:journals/corr/abs-2302-06590,DBLP:journals/corr/abs-2402-01580}. Beyond text-based tasks, the focus has expanded to multimodal models that process and generate not only text but also images, achieving significant success in domains such as visual question answering~\citep{DBLP:conf/icml/RadfordKHRGASAM21} and text-to-image generation~\citep{DBLP:conf/icml/RameshPGGVRCS21}.

Despite these successes, the performance of large models on tasks that require a combination of skills remains unclear. Real-world tasks often demand a blend of skills. For example, a typical task like ``navigating to the kitchen to fetch ten apples'' involves spatial reasoning to understand the environment and plan a path around obstacles, together with basic arithmetic to ensure that exactly ten apples are retrieved. This example illustrates the multifaceted nature of real-world tasks. While various benchmarks focus on specific skills~\citep{DBLP:journals/corr/abs-2107-03374,DBLP:conf/nips/HendrycksBKABTS21,DBLP:conf/iclr/HendrycksBBZMSS21,DBLP:conf/acl/LinHE22}, there is a lack of benchmarks evaluating how large models perform on tasks that require a combination of different skills.

\begin{figure*}[t!]
  \centering
  \begin{minipage}[t]{0.185\linewidth}
    \centering
    \begin{subfigure}[b]{\linewidth}
      \caption*{\scriptsize \textbf{Task 28}: Collect all red shapes without standing on the color green.}
      \includegraphics[width=\textwidth]{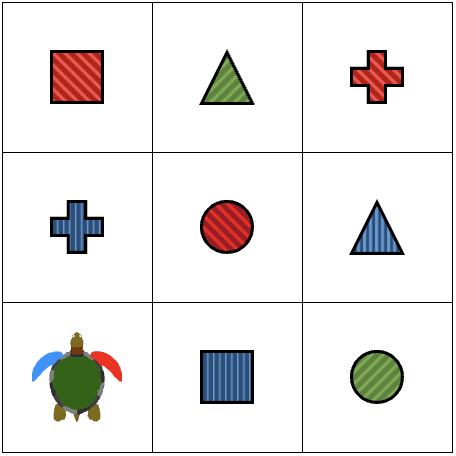}
      \vspace{-4pt}
    \end{subfigure}
    \begin{subfigure}[b]{\linewidth}
      \textbf{\scriptsize Required Skills:} \scriptsize{Logic, Basic Actions}\\
    \end{subfigure}
    \begin{subfigure}[b]{\linewidth}
      \textbf{\scriptsize Solution Code:}
      \begin{lstlisting}
def Run():
  $\fd{}$
  $\fd{}$
  $\bk{}$
  $\rt{}$
  $\fd{}$
  $\fd{}$
  $\lt{}$
  $\fd{}$
  $$
  $$
  $$
      \end{lstlisting} 
    \end{subfigure}
  \end{minipage}
  \hspace{0.005\linewidth}
  \begin{minipage}[t]{0.185\linewidth}
    \centering
    \begin{subfigure}[b]{\linewidth}
      \caption*{\scriptsize \textbf{Task 38}: Collect exactly 10 strawberries.\\}
      \includegraphics[width=1\textwidth]{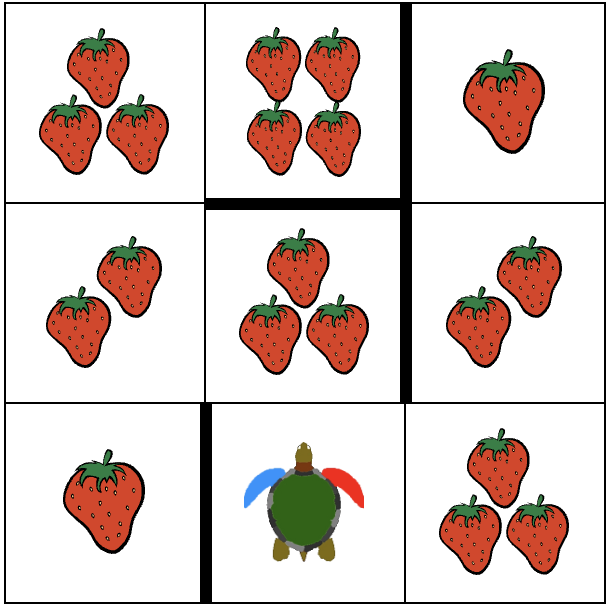}
      \vspace{-4pt}
    \end{subfigure}
    \begin{subfigure}[b]{\linewidth}
      \textbf{\scriptsize Required Skills:} \scriptsize{Math, Basic Actions}\\
    \end{subfigure}
    \begin{subfigure}[b]{\linewidth}
      \textbf{\scriptsize Solution Code:}
      \begin{lstlisting}
def Run():
  $\fd{}$
  $\lt{}$
  $\fd{}$
  $\bk{}$
  $\lt{}$
  $\fd{}$
  $\lt{}$
  $\fd{}$
  $\lt{}$
  $\fd{}$
  $$
      \end{lstlisting} 
    \end{subfigure}
  \end{minipage}
  \hspace{0.005\linewidth}
  \begin{minipage}[t]{0.185\linewidth}
    \centering
    \begin{subfigure}[b]{\linewidth}
      \centering
      \caption*{\scriptsize \textbf{Task 65}: Draw the picture using the colors yellow, green, blue and red.\\}
      \includegraphics[width=0.87\textwidth]{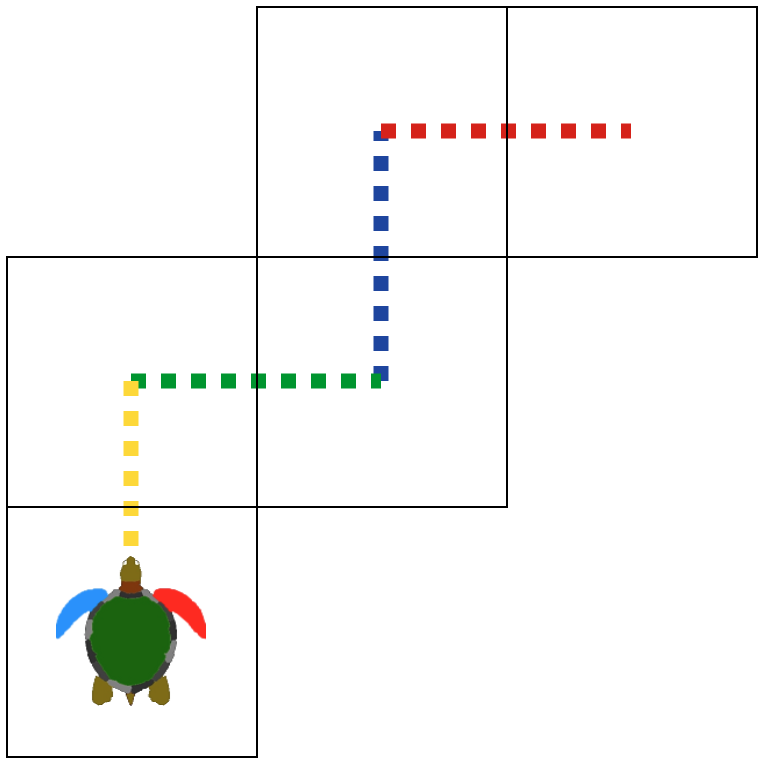}
      \vspace{13pt}
    \end{subfigure}
        \begin{subfigure}[b]{\linewidth}
          \textbf{\scriptsize Required Skills:} \scriptsize{Draw, Variables, Basic Actions}\\
        \end{subfigure}
        \begin{subfigure}[b]{\linewidth}
          \textbf{\scriptsize Solution Code:}
          \begin{lstlisting}
def Run():
  $\setpc{}$("yellow")
  $\fd{}$
  $\rt{}$
  $\setpc{}$("green")
  $\fd{}$
  $\lt{}$
  $\setpc{}$("blue")
  $\fd{}$
  $\rt{}$
  $\setpc{}$("red")
  $\fd{}$
          \end{lstlisting} 
    \end{subfigure}
  \end{minipage}
  \hspace{0.005\linewidth}
  \begin{minipage}[t]{0.185\linewidth}
    \centering
    \begin{subfigure}[b]{\linewidth}
      \caption*{\scriptsize \textbf{Task 73}: Draw the picture in green. \\}
      \includegraphics[width=\textwidth]{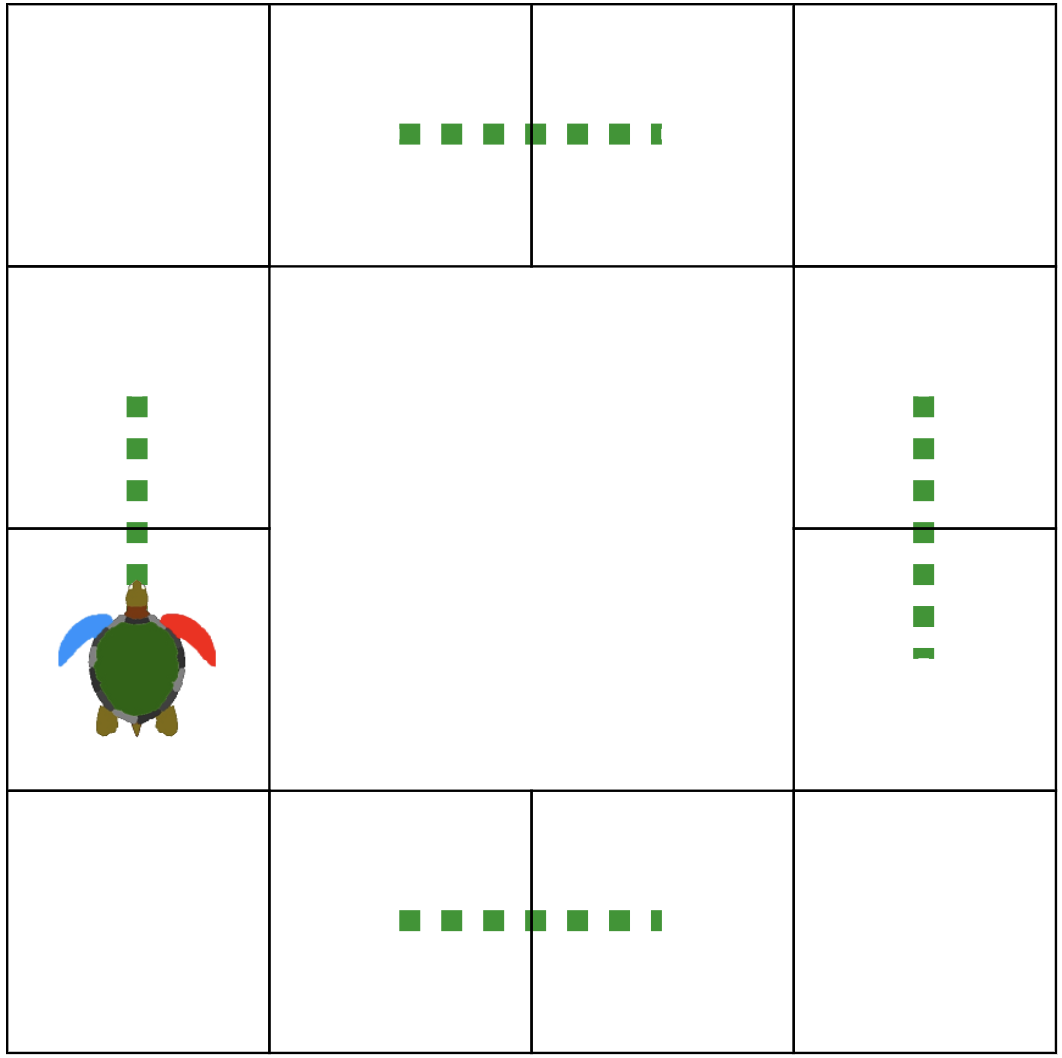}
      \vspace{-4pt}
    \end{subfigure}
        \begin{subfigure}[b]{\linewidth}
          \textbf{\scriptsize Required Skills:} \scriptsize{Draw, Variables, Loops}\\
        \end{subfigure}
        \begin{subfigure}[b]{\linewidth}
          \textbf{\scriptsize Solution Code:}
          \begin{lstlisting}
def Run():
  $\Repeat{}$ i in range(4):
    $\setpc{}$("green")
    $\fd{}$
    $\setpc{}$("white")
    $\fd{}$
    $\rt{}$
    $\fd{}$
    $$
    $$
    $$
    $$
          \end{lstlisting} 
        \end{subfigure}
  \end{minipage}
  \hspace{0.005\linewidth}
  \begin{minipage}[t]{0.185\linewidth}
    \centering
    \begin{subfigure}[b]{\linewidth}
      \caption*{\scriptsize \textbf{Task 87}: Find the strawberry with just 6 commands.  \\ \\ }
      \includegraphics[width=1.03\textwidth]{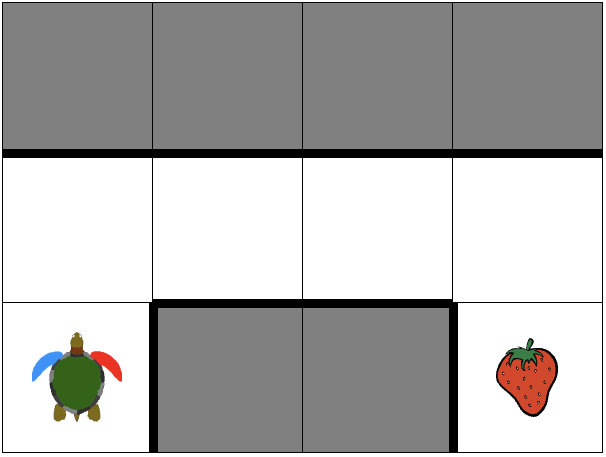}
      \vspace{7pt}
    \end{subfigure}
        \begin{subfigure}[b]{\linewidth}
          \textbf{\scriptsize Required Skills:} \scriptsize{Code Constraints, Loops}\\
        \end{subfigure}
        \begin{subfigure}[b]{\linewidth}
          \textbf{\scriptsize Solution Code:}
          \begin{lstlisting}
def Run():
  $\fd{}$
  $\rt{}$
  $\Repeat{}$ i in range(3):
    $\fd{}$
  $\rt{}$
  $\fd{}$
  $$
  $$
  $$
  $$
  $$
          \end{lstlisting} 
        \end{subfigure}
  \end{minipage}
  \caption{Examples of real-world tasks, required skills, and solution codes in \platformMini{}.}
  \label{fig.example_tasks}
  \vspace{-10pt}
\end{figure*}

\looseness-1To bridge this gap, we introduce \benchmark{}, a benchmark for program synthesis in the visual programming domain. Our benchmark is constructed using the Mini-level of the XLogoOnline platform~\citep{xlogoonline}, featuring $85$ real-world and $1,000$ synthetic visual programming tasks, each demanding a blend of diverse skills. Figure~\ref{fig.example_tasks} illustrates examples of these tasks. Each task includes a visual grid with a turtle that needs to be directed to complete a specific goal. For example, in Task 28, the goal is to direct the turtle to collect all red shapes without stepping on the color green, requiring logical reasoning, spatial reasoning, planning, and basic programming skills. Task 38 requires additional math word problem-solving to collect $10$ strawberries. These tasks provide a testbed for evaluating how large models perform on tasks that require a combination of skills, presenting a unique challenge to current large models.

We evaluate large models on these tasks and find that GPT-4V and Llama3-70B struggle, with success rates of only $20\%$ and $2.35\%$, respectively. Reasoning model DeepSeek-R1-Distill-Llama3-70B performs better, achieving $44.71\%$ success rate. However, a significant gap remains compared to human performance, as these tasks are designed for students up to 2nd grade, where humans can successfully solve almost all tasks~\cite{DBLP:journals/eatcs/Staub21}.
This indicates that current large models are not yet capable of effectively solving visual programming tasks that require a combination of skills. Figure~\ref{fig.intro.radar_real_skills} compares the performance of large models across different skill dimensions on these tasks. To improve performance, we develop a fine-tuning pipeline by leveraging a large-scale synthetic dataset containing over $80,000$ visual programming tasks. Our fine-tuned Llama3-8B model outperforms GPT-4V, Llama3-70B, and DeepSeek-R1-Distill-Llama3-70B, achieving a $54.12\%$ success rate. Moreover, we leverage emulator feedback to design a curriculum over the training data distribution, improving performance by $6.1\%$ over standard supervised fine-tuning. 

\looseness-1
Our contributions are as follows: 
First, we introduce \benchmark{}, a program synthesis benchmark based on the \platformAll{} platform to evaluate large models in visual programming, which requires a blend of skills. 
Second, we develop a fine-tuning pipeline that includes synthetic dataset generation and supervised fine-tuning, along with an emulator-driven fine-tuning technique that improves standard supervised fine-tuning performance by $6.1\%$.
Third, we conduct extensive experiments to benchmark the performance of different models, providing an in-depth failure analysis and a detailed analysis of their expertise across multiple skill dimensions. 
We will publicly release the benchmark for future research on program synthesis in visual programming.\footnote{\url{https://github.com/machine-teaching-group/acl2025-xlogominiprog}}



\section{Related Work}\label{sec.related-work}

\paragraph{Multimodal benchmarks for large models.}
\looseness-1
Existing works have developed many multimodal benchmarks to evaluate the visual understanding and reasoning capabilities of multimodal models~\citep{DBLP:conf/iclr/HendrycksBBZMSS21,DBLP:conf/icml/YuYLWL0WW24,DBLP:journals/corr/abs-2405-02287,DBLP:conf/nips/DevlinBSHK17}. Our work falls into this broader category but focuses on program synthesis in visual programming domains that demand both visual reasoning and programming skills. Furthermore, our benchmark is built on a widely used educational programming platform, providing practical value for AI-assisted programming education.

\paragraph{Program synthesis benchmarks for large models.} 
Program synthesis aims to automatically generate programs from specifications. Recently, numerous recent works have focused on training large models specifically for program synthesis~\citep{DBLP:journals/corr/abs-2107-03374,DBLP:journals/corr/abs-2308-12950,DBLP:conf/iclr/FriedAL0WSZYZL23,DBLP:conf/iclr/NijkampPHTWZSX23}. To evaluate these large models, many program synthesis benchmarks have been developed, such as HumanEval~\citep{DBLP:journals/corr/abs-2107-03374}, MBPP~\citep{DBLP:journals/corr/abs-2108-07732}, and APPS~\citep{DBLP:conf/nips/HendrycksBKMAGB21}. However, these benchmarks focus on generating code from natural language or docstrings for general programming languages such as Python~\citep{DBLP:journals/corr/abs-2107-03374,DBLP:journals/corr/abs-2108-07732,DBLP:conf/nips/HendrycksBKMAGB21}. Our benchmark focuses on program synthesis in the visual programming domain. While our benchmark covers basic programming like loops and variables, it requires models to combine spatial, logical, and programming skills, posing unique challenges not addressed by these program synthesis benchmarks.

\paragraph{Large models for visual programming.}
Visual programming has been studied in various scenarios, such as task synthesis~\citep{DBLP:conf/nips/AhmedCEFGRS20,DBLP:conf/aied/GhoshTDS22,chao2024xlogo,puadurean2023neural}, program synthesis~\citep{DBLP:conf/iclr/BunelHDSK18,DBLP:conf/iclr/ChenLS19}, and student modeling~\citep{DBLP:journals/corr/abs-2310-10690}. With the rise of large models, some initial works evaluate ChatGPT~\citep{chatgpt} and GPT-4~\citep{gpt4} in these scenarios, showing that large models struggle with visual programming tasks~\citep{puadurean2023neural,DBLP:journals/corr/abs-2310-10690,DBLP:conf/icer/Singla22}. In contrast, we provide a comprehensive benchmark that evaluates a broader range of models and skills for program synthesis in visual programming.



\begin{figure*}[t!]
    \centering
    \begin{subfigure}[b]{1\textwidth}
      \centering
      \scalebox{0.81}{
      \begin{tabular}{p{2cm}r|p{3cm}r|p{3cm}r|p{3cm}r|p{1.8cm}r}
      \toprule
      \textbf{Task Type} & \# & \textbf{Code Constraints} & \# & \textbf{Code Concepts}  & \# & \textbf{Code Length} & \# & \textbf{Grid Size} & \# \\
      \midrule
      \find & $33$ & \none & $54$ & \ProgBasic{} & $47$  & Short (1-5) & $41$ & $Size \leq 3$ & $59$ \\
      \draw & $33$ & \atmost & $21$ & \ProgBasicRepeat{} & $24$ & Medium (6-10) & $29$ & $Size = 4 $ & $15$ \\
      \counting & $10$ & \exactly & $6$ & \ProgBasicSetpc{} & $7$ & Long (11-17)  & $15$ & $ Size=5$ & $4$ \\
      \logic & $9$ & \startby{} & $4$ & \ProgAll{} & $7$ & & & $ Size=6$ & $4$ \\
      & & \hybrid{} & $0$ & & & &  & $ Size\geq 7$ & $3$ \\
      \midrule
      {Total} & $85$ & {Total} & $85$ & {Total} & $85$ & {Total} & $85$ & {Total} & $85$ \\
      \bottomrule
      \end{tabular}
      }
      \caption{Task distribution of \benchmarkreal{} dataset.}
      \label{tab.real_dataset_stats}
      \vspace{0.5em}
    \end{subfigure}
    \begin{subfigure}[b]{0.46\textwidth}
      \centering
      \includegraphics[width=\linewidth]{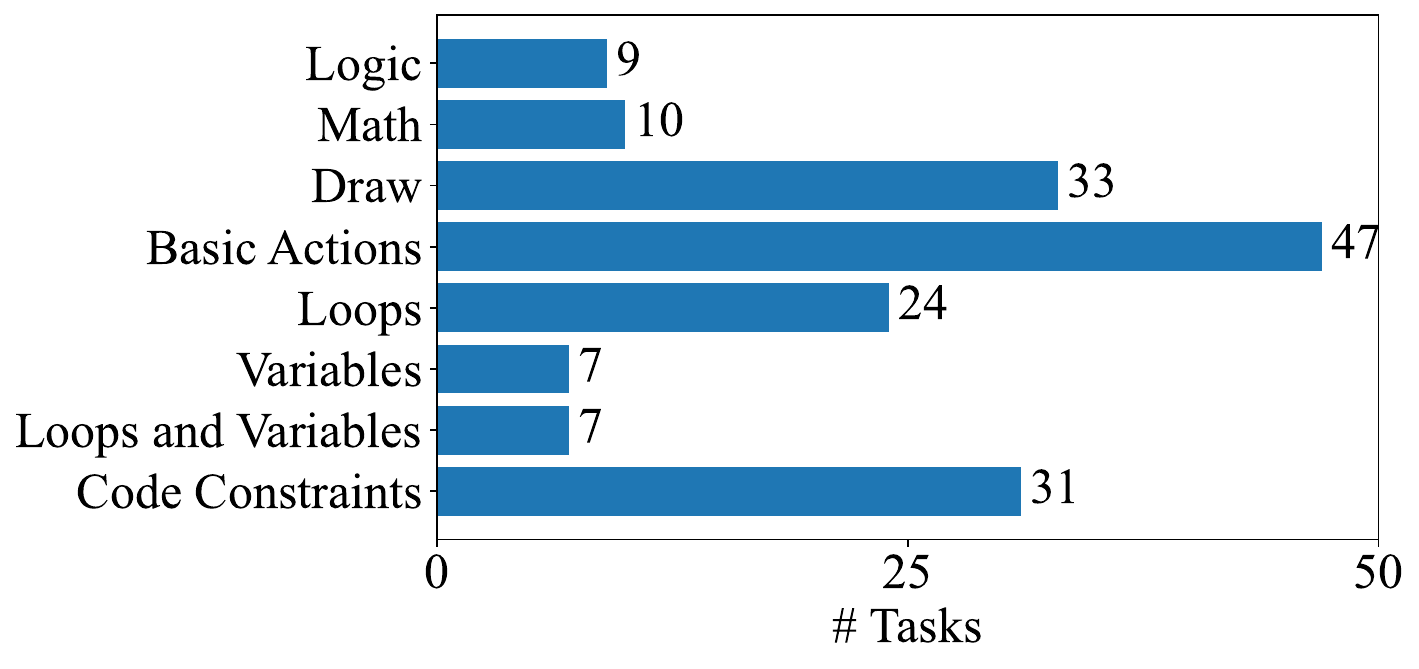}
      \caption{Skill distribution of \benchmarkreal{} dataset.}
      \label{fig.hist_real_distribution}
    \end{subfigure}
    \begin{subfigure}[b]{0.525\textwidth}
      \begin{lstlisting}[frame=single, breaklines=true, basicstyle=\ttfamily\scriptsize,belowskip=7mm,xleftmargin=2em,xrightmargin=1em,framesep=2pt]
code C    := def Run() Do b

rule b    := a | b; b | repeat(x) do b 

action a  := forward | backward | left | right | setpc(r)

color r   := red | blue | green | white | black | yellow

iter x    := 2 | 3 | 4 | 5 | 6 | 7 | 8 | 9 | 10 
      \end{lstlisting}
      \caption{Code DSL.}
      \label{fig.codedsl}
    \end{subfigure}
    \caption{
    Statistics of the \benchmarkreal{} dataset and the code DSL.
    (a) shows the task distribution across five dimensions within \benchmarkreal{}.
    (b) illustrates the skill distribution. To describe these skills, we extract various aspects from task type, code constraints, and code concepts as detailed in (a), and consolidated these aspects into broader categories, which we refer to as \emph{skills}. A task may require multiple skills (see Figure~\ref{fig.example_tasks}).
    (c)~shows the code DSL used in the \platformMini{} domain.}
    \label{fig.real_dataset_statistics}
  \end{figure*}

\section{Background and Synthesis Objective} \label{sec.background}

In this section, we provide background on the \platformAll{} visual programming platform and then introduce the program synthesis objective.

\subsection{Background on \platformMini{}}

\looseness-1
XLogoOnline~\citep{xlogoonline} is a visual programming platform based on the Logo programming language~\citep{pea1987logo} and is widely used by tens of thousands of students every year~\citep{DBLP:conf/issep/HromkovicSS17,DBLP:journals/eatcs/Staub21}. In this work, we focus on the Mini-level (\platformMini{}).
In \platformMini{}, each task includes a text description of the goal and code constraints, along with a two-dimensional visual grid. The visual grid features a turtle and various elements such as fruits, shapes, colors, lines, walls, and forbidden areas. To solve the task, one needs to write a program to direct the turtle's movement in the visual grid to achieve the specified goal. Figure~\ref{fig.example_tasks} shows illustrative examples of tasks, the required skills, and solution codes.

\paragraph{Required skills for \platformMini{}.} 
We examine the skills required for solving visual programming tasks in \platformMini{}. Specifically, the visual programming tasks in our domain cover the following skills:
(i)~\textit{Logic}: Understanding underlying logical relationships specified in the goal; (ii)~\textit{Math}: Applying basic arithmetic to solve the task; (iii)~\textit{Draw}: Identifying patterns and generating the corresponding code; (iv)~\textit{Basic actions}: Moving and changing directions using only basic commands; (v)~\textit{Loops}: Utilizing loops to repeat commands multiple times; (vi)~\textit{Variables}: Using variables to set and update colors to draw lines with a specific color; (vii)~\textit{Loops and Variables}: Integrating loops with variables to solve a task; (viii)~\textit{Code Constraints}: Adhering to specific code constraints such as maximum code length.

\subsection{Program Synthesis Objective} \label{sec.formulation}

Next, we formally define task and code specifications and introduce our synthesis objective.

\paragraph{Task specifications.} In \platformMini{}, a task $\task := (\goal, \cons, \grid)$ consists of a goal $\goal$, code constraints $\cons$, and a visual grid world $\grid$. 
The goal $\goal$ defines the turtle's objective. The code constraints $\cons$ specify the requirements for a solution code. There are five types of code constraints: \none{} (no restrictions), \atmost{} (maximum number of commands), \exactly{} (exact number of commands), \startby{} (initial command sequence), and \hybrid{} (combination of constraints). 
The visual grid world $\grid$ is a two-dimensional visual grid featuring a turtle and various elements. We define the grid size as the maximum dimension of the grid. For example, in Figure~\ref{fig.example_tasks} (Task 87), the goal is ``Find the strawberry'', the code constraint is ``use just 6 commands'' (\atmost), and the visual grid world depicts a $3 \times 4$ grid ($size=4$) with a turtle, a strawberry, and forbidden areas marked by gray cells.

\paragraph{Code specifications.} The code space of \platformMini{} is defined by the domain-specific language (DSL) depicted in Figure~\ref{fig.codedsl}. Note that while the DSL defines the formal structure and syntax, we implement it using a Python-style code representation to leverage large models' pre-trained knowledge of Python programming. A \emph{solution code} for a given task is the code that meets the task's code constraints and achieves the specified goal when executed in the visual grid world. In Figure~\ref{fig.example_tasks}, a solution code is provided below each task.

\paragraph{Program synthesis objective.} \label{sec.synthesis_objective}
Our objective is to develop a synthesizer function, $f: \task \rightarrow \code$, which generates a solution code $\code$ for a given visual programming task $\task$.
To evaluate $f$ on a task $\task$, we first use $f$ to synthesize a code $\hat{\code}$ and then execute the synthesized code $\hat{\code}$ using an emulator. The emulator outputs \emph{success} if the synthesized code $\hat{\code}$ successfully solves the task $\task$ and adheres to code constraints; otherwise, the emulator outputs \emph{fail}. We use \emph{success} as the main evaluation metric.
Given a dataset $\mathcal{D}_{\text{eval}} = \{\task_i\}_{i=1}^N$, we calculate the success rate of $f$ on this dataset as the overall performance. 
We curate a dataset $\benchmarkreal{}$ of $N=85$ real-world visual programming tasks from \platformMini{}, and we use this as one of the main datasets for evaluation. In Figures~\ref{tab.real_dataset_stats} and~\ref{fig.hist_real_distribution}, we show the overall distribution of this dataset and the number of tasks requiring specific skills, respectively.



\section{Methodology for Synthetic Dataset Generation and Fine-tuning}

As discussed in Section~\ref{sec.intro}, existing large models such as GPT-4V and Llama3-70B struggle with visual programming tasks in \platformMini{}. To address this, we develop a two-stage fine-tuning pipeline consisting of synthetic dataset generation and supervised fine-tuning. This section details the dataset generation process and the methodology for fine-tuning large models on the synthetic dataset.



\begin{figure*}[t!]
    \centering
    \begin{subfigure}[b]{\textwidth}
        \centering
        \begin{minipage}[c]{1\linewidth}
            \centering
            \scalebox{0.70}{
            \begin{tabular}{p{2cm}r|p{3cm}r|p{3cm}r|p{2.5cm}r|p{2.5cm}r}
            \toprule
            \textbf{Task Type} & \# & \textbf{Code Constraints} & \# & \textbf{Code Concepts}  & \# & \textbf{Code Length} & \# & \textbf{Grid Size} & \# \\
            \midrule
            \find & $36,055$ & \none & $34,680$ & \ProgBasic{} & $53,779$  & Short (1-5) & $20,985$ & $Size \leq 3$ & $35,908$ \\
            \draw & $24,851$ & \atmost & $29,354$ & \ProgBasicRepeat{} & $24,432$ & Medium (6-10)  & $45,682$ & $Size = 4 $ & $25,933$ \\
            \counting & $14,994$ & \exactly & $16,169$ & \ProgBasicSetpc{} & $5,931$ & Long (11-17) & $22,386$ & $ Size=5$ & $14,852$ \\
            \logic & $13,153$ & \startby{} & $1,430$ & \ProgAll{} & $4,911$ & & & $ Size=6$ & $8,061$ \\
            & & \Hybrid{} & $7,420$ & & & &  & $ Size\geq 7$ & $4,299$ \\
            \midrule
            {Total} & $89,053$ & {Total} & $89,053$ & {Total} & $89,053$ & {Total} & $89,053$ & {Total} & $89,053$ \\
            \bottomrule
            \end{tabular}
            }
        \end{minipage}%
        \caption{Task distribution of \benchmarksim{} dataset.}
        \label{tab.dataset_stats}
        \vspace{0.5em}
    \end{subfigure}
    \begin{subfigure}[b]{0.52\textwidth}
        \centering
        \includegraphics[width=\linewidth]{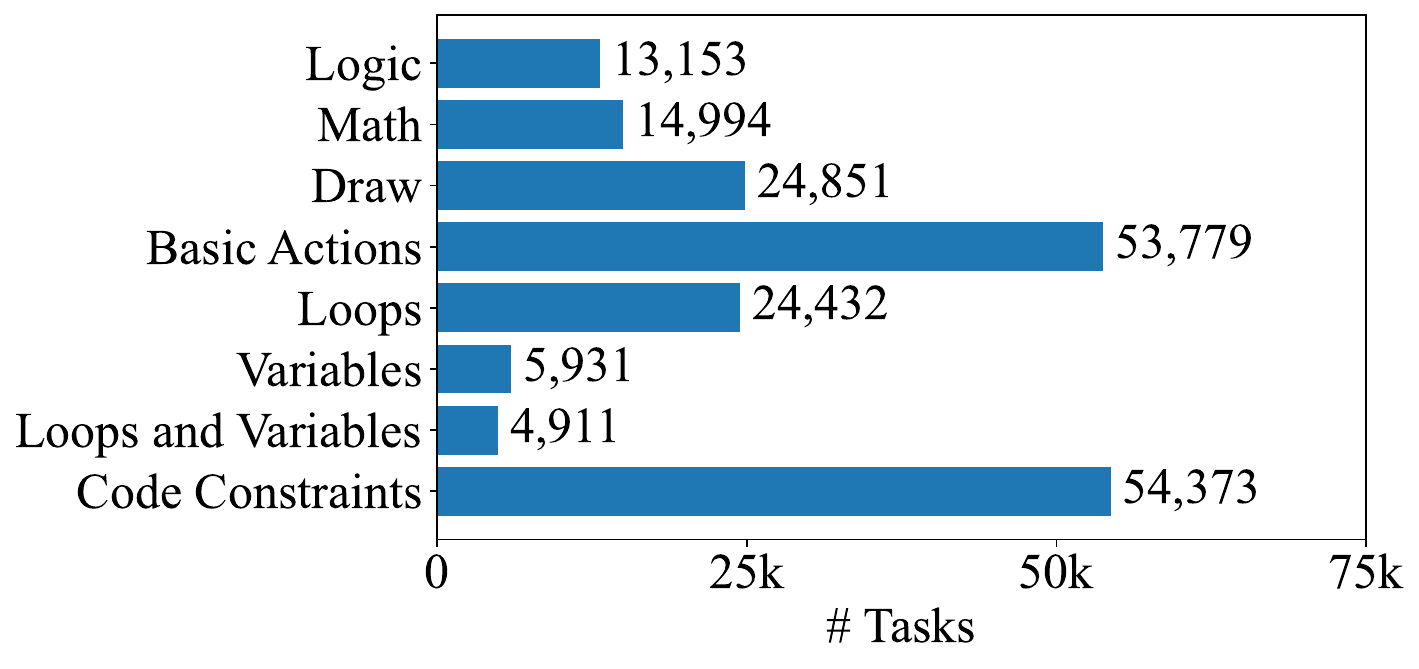}
        \caption{Skill distribution of \benchmarksim{} dataset.}
        \label{fig.syn_skill_distribution}
    \end{subfigure}
    \hfill
    \begin{subfigure}[b]{0.47\textwidth}
        \centering
        \begin{tabular}{lrr}
            \toprule
            Dataset & Purpose & \# \\
            \midrule
            \benchmarksimtrain{} & Train & $87,053$ \\
            \benchmarksimval{} & Validation & $1,000$ \\
            \benchmarksimeval{} & Evaluation & $1,000$ \\
            \midrule
            \benchmarksim{} &  & $89,053$ \\
            \bottomrule
        \end{tabular}
        \vspace{2.0em}
        \caption{Dataset split of \benchmarksim{} dataset.}
        \label{tab.dataset_split}
    \end{subfigure}
    \caption{Statistics of the synthetic \benchmarksim{} dataset. (a) and (b) show the task distribution and the skill distribution, respectively. 
    (c) shows the dataset split. 
     }
    \label{fig.syn_dataset_statistics}
\end{figure*}

\subsection{Synthetic Dataset Generation} \label{sec.dataset_generation}

\looseness-1Our goal is to develop a large synthetic dataset to train models~\citep{DBLP:conf/iclr/BunelHDSK18}. To achieve this, we adopt the task synthesis techniques from~\citep{DBLP:conf/nips/AhmedCEFGRS20,chao2024xlogo}, which were developed to automatically generate high-quality tasks in visual programming education domains. Instead of random task generation, these techniques employ symbolic execution and constraint satisfaction, enabling more controlled and systematic task synthesis, such as specifying task types, code concepts, and code lengths.
However, we further adapt these techniques to generate a larger, more diverse dataset for model training (see Appendix~\ref{sec:appendix:synthetic_dataset_generation} for more details).

\paragraph{Dataset generation process and statistics.} 
\looseness-1We use the adapted task synthesis technique to generate a synthetic dataset as follows: 
(i) We manually craft a solution code for each task in the \benchmarkreal{} dataset, resulting in a set $\{(\task_i, \code_i)\}_{i=1}^{85}$;
(ii) For each $(\task_i, \code_i)$, we generate up to $1,500$ synthesized tasks and their solution codes.
To ensure the quality of the dataset, we take the following processing steps: we remove any duplicate task-code pairs to maintain diversity, conduct a correctness check on the generated solution codes using the emulator, and exclude any task-code pairs present in the real-world \benchmarkreal{} dataset from our synthetic dataset. This last processing step guarantees that the model has not seen any tasks from the evaluation dataset during training. 
We ultimately produce the synthetic dataset \benchmarksim{} with $89,053$ task-code pairs. The statistics of this dataset are detailed in Figure~\ref{fig.syn_dataset_statistics}. 
From this synthetic dataset, we randomly select $1,000$ samples for validation, $1,000$ samples for evaluation, and use the remaining samples for training. We use this synthetic evaluation dataset ($1,000$ samples), referred to as \benchmarksimeval{}, to complement the real-world dataset \benchmarkreal{} in evaluating the model's performance. We provide full details of the dataset generation process and dataset quality assessment in Appendix~\ref{sec:appendix:dataset}. 

\subsection{Methodology for Fine-tuning} \label{sec.finetuning}

\paragraph{Supervised fine-tuning.} 
Supervised fine-tuning is commonly used to improve large models' performance on domain tasks. In our domain, one can use the synthetic dataset \benchmarksim{} to fine-tune large models, where the model receives a natural language task description as input and outputs Python-style code. The model is optimized to minimize the cross-entropy loss between the predicted code and the ground truth solution code.



\begin{figure*}[t]
    \centering
    \begin{subfigure}[b]{0.38\linewidth}
        \centering
        \scalebox{1}{
            \setlength\tabcolsep{5pt}
            \renewcommand{\arraystretch}{1.2}
\footnotesize
\begin{tabular}{|p{\linewidth}|}
    \hline
    \\[-8pt]
    You are presented with a visual programming task involving a goal, a grid, a turtle, various items (or lines). You need to write Python code that enables the turtle to accomplish the goal within the grid.
    \\\\
    \promptinput{\{description\_of\_grid\_properties\}}
    \\
    \promptinput{\{description\_of\_python\_functions\}}
    \\
    Now, write a correct Python code that successfully solves the following task.
    \\\\
    \#\#\# Task:\\
    \promptinput{\{description\_of\_task\}}\\

    \#\#\# Goal:\\
    \promptinput{\{description\_of\_goal\}}\\

    \#\#\# Correct code:
    \\
    \hline
\end{tabular}

        }
        \caption{Prompt template.}
        \label{fig.prompt_template_nl}
    \end{subfigure}
    \hfill
    \begin{subfigure}[b]{0.56\linewidth}
        \centering
        \begin{subfigure}[b]{\linewidth}
            \centering
            \includegraphics[width=0.895\linewidth]{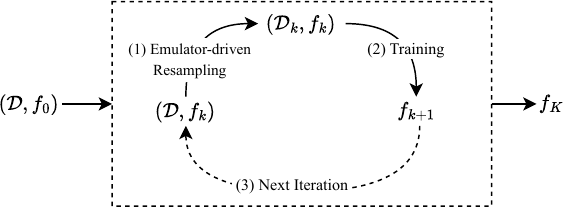}
            \caption{Overview of emulator-driven fine-tuning.}
            \label{fig.technique_overview}
        \end{subfigure}
        \begin{subfigure}[b]{\linewidth}
            \centering
            \vspace{0.5em}
            \includegraphics[width=0.90\linewidth]{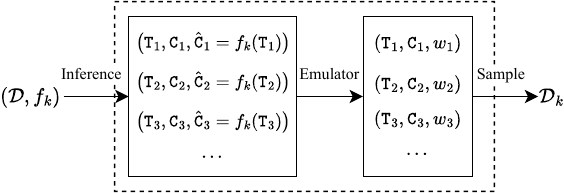}
            \caption{Emulator-driven resampling.}
            \label{fig.technique_resample}
        \end{subfigure}
    \end{subfigure}
    \caption{(a)~shows the prompt template for fine-tuning. This prompt has several \promptinput{placeholders} to include details for the descriptions of different aspects of the task. More details can be found in the Appendix. 
    (b)~provides an overview of emulator-driven fine-tuning, starting with the dataset $\mathcal{D}$ and initial model $f_0$, and iteratively resampling and training to produce the final model $f_K$.
    (c)~illustrates the resampling process in emulator-driven fine-tuning to create the dataset $\mathcal{D}_k$.}
    \label{fig.technique}
    \vspace{-1em}
\end{figure*}

\paragraph{Emulator-driven feedback for fine-tuning.} \label{sec.emulator}

Standard supervised fine-tuning assigns equal weights to all training samples. However, our domain tasks require varying skills and have different difficulty levels (see Figure~\ref{fig.syn_dataset_statistics}). Moreover, some skills are prerequisites for mastering advanced ones. For instance, a model typically needs to understand basic actions before mastering loops and variables, and it solves tasks with shorter code lengths before tackling longer ones. Thus, treating all tasks with equal importance can be suboptimal in our setting~\citep{DBLP:conf/icml/BengioLCW09}. 
To address this, we introduce emulator-driven fine-tuning, which designs a curriculum over the training data distribution by leveraging emulator feedback. The key idea is to dynamically adjust the data distribution based on emulator evaluation, assigning higher weights to challenging tasks and progressively guiding the model from simpler to more complex problems.
The overall process is shown in Figures~\ref{fig.technique_overview} and~\ref{fig.technique_resample}. Given an initial model $ f_0 $ and the training dataset $ \mathcal{D} $, our goal is to learn a final model $ f_K $. To achieve this, at each training epoch $k$, we first perform the \emph{emulator-driven resampling} step (see Figure~\ref{fig.technique_resample}), where model $ f_k $ infers on the training dataset $ \mathcal{D} $ to obtain the predicted code $ \hat{\code}_i $ for each task $ \task_i $. Then, we evaluate each predicted code using an emulator and update the weight $ w_i $ for $ (\task_i, \code_i) $ as follows:
\begin{equation}
w_i = \frac{1}{|\mathcal{D}|}\big[1 + \beta \cdot \mathbb{I}\big(\text{Emulator}(\task_i, \hat{\code}_i) = \emph{fail}\big)\big],
\label{eq.weight}
\end{equation}
where $ \mathbb{I}(\cdot) $ is an indicator function that returns $1$ if the predicted code fails to solve $ \task_i $, and $0$ otherwise. The hyperparameter $\beta$ is adjustable, with a larger $\beta$ encouraging the model to focus more on its mistakes, and $\beta=0$ is equivalent to fine-tuning without resampling. Then, we sample from the training dataset $ \mathcal{D} $ according to the categorical distribution $ w_i' = w_i/{{\scriptstyle \sum_{j=1}^{|\mathcal{D}|}} w_j} $ to obtain a resampled dataset $ \mathcal{D}_k $. After resampling, we perform the \emph{training} step, where we train model $ f_k $ on $\mathcal{D}_k$ to obtain $f_{k+1}$. We repeat the resampling and training steps until the model converges or reaches a predefined number of training epochs, yielding the final model $ f_K $. To reduce computational costs, resampling can be performed at fixed intervals (set to $3$ epochs in our experiments).



\section{Experimental Evaluations}
\label{sec.exp}

In this section, we present a comprehensive evaluation of large models on \benchmark{}. We begin by describing the experimental setup in Section~\ref{sec.exp_setup}, followed by the main results in Section~\ref{sec.exp_results}. Next, we provide a comparative analysis in Section~\ref{sec.exp_ablations} and a failure analysis in Section~\ref{sec.failure_analysis}.



\begin{figure*}[ht!]
  \centering
  \scalebox{0.73}{
  \begin{tabular}{lrrrrrr}
    \toprule
    & \multicolumn{3}{c}{\benchmarkreal{} ($85$ tasks)} & \multicolumn{3}{c}{\benchmarksimeval{} ($1,000$ tasks) } \\
    \cmidrule(lr){2-4} \cmidrule(lr){5-7} 
    & Format (\%) & No-Crash (\%) & Success (\%) & Format (\%) & No-Crash (\%) &  Success (\%)\\
    \midrule
    \textbf{\emph{Base LLMs (text-only):}} &  &   &   &   &   &  \\
    \quad GPT-3.5 {\small (\texttt{gpt-3.5-turbo-0125})} & $92.94$ &   $11.76$   & $\ 1.18$  & $87.60$  & $\ 9.50$  & $1.60$ \\
    \quad GPT-4 {\small (\texttt{gpt-4-turbo-2024-04-09})} & $\textbf{95.29}$ & $\textbf{38.83}$  &  $\textbf{12.94}$ & $\textbf{97.40}$  & $\textbf{16.80}$  & $\textbf{5.30}$ \\
    \quad Llama3-8B &  $48.24$   &  $\ 5.88$  & $\ 0.00$ & $40.90$ & $\ 2.80$ & $0.60$ \\
    \quad Llama3-70B &   $67.06$  &  $\ 8.24$ &  $\ 2.35$ & $15.50$ & $\ 1.20$ & $0.30$ \\
    \quad Llama2-7B & $27.06$  & $\ 5.88$  & $\ 0.00$  & $21.90$  & $\ 2.90$  & $0.40$ \\
    \quad Llama2-13B &  $60.00$ & $\ 7.06$  &  $\ 0.00$ & $54.40$  & $\ 3.50$  & $0.40$ \\
    \quad Llama2-70B & $28.24$  & $\ 7.06$  & $\ 0.00$  & $38.30$  & $\ 1.10$  & $0.10$ \\

    \midrule
    \textbf{\emph{Base VLMs (text + vision):}} &  &   &   &   &   &  \\
    \quad GPT-4o {\small (\texttt{gpt-4o-2024-11-20})} & $\mathbf{100.00}$ & $36.47$ & $\mathbf{22.35}$ & $\mathbf{99.10}$ & $\mathbf{18.30}$ & $\mathbf{5.90}$ \\
    \quad GPT-4V {\small (\texttt{gpt-4-turbo-2024-04-09})}  & $96.47$ & $\textbf{47.06}$ & $20.00$ & $95.50$ & $18.10$ & $5.50$ \\
    \quad Llava1.5-7B & $10.59$ & $1.18$ & $0.00$ & $3.20$ & $0.00$ & $0.00$ \\
    \quad Llava1.5-13B & $10.59$ & $2.35$  &  $0.00$ & $9.00$  & $2.10$  & $0.00$ \\
    \quad InternVL2-8B & $0.00$ & $0.00$ & $0.00$ & $56.90$ & $3.80$ & $0.00$ \\
    \quad InternVL2-Llama3-76B & $77.65$ & $31.76$ & $9.41$ & $40.50$ & $6.10$ & $1.50$ \\
    \quad Qwen2VL-7B & $43.53$ & $9.41$ & $0.00$ & $14.30$ & $2.10$ & $0.20$ \\
    \quad Qwen2VL-72B & $28.24$ & $11.76$ & $0.00$ & $36.50$ & $4.40$ & $0.40$ \\
    \quad NVLM-D-72B & $61.18$ & $8.24$ & $1.18$ & $67.40$ & $8.30$ & $2.00$ \\
    \quad Molmo-7B-D & $75.29$ & $8.24$ & $0.00$ & $66.00$ & $7.70$ & $0.60$ \\
    \quad Molmo-72B & $4.71$ & $1.18$ & $1.18$ & $6.40$ & $0.70$ & $0.40$ \\
    \midrule
    \textbf{\emph{Reasoning LLMs (text-only):}} &  &   &   &   &   &  \\
    \quad o1 ({\small \texttt{o1-2024-11-12}}) &  $\textbf{100.00}$ & $\textbf{97.65}$ & $\textbf{76.47}$ & $\textbf{99.08}$ & $\textbf{47.38}$ & $23.18$ \\
    \quad DeepSeek-R1-Distill-Llama-8B & $38.82$ & $21.18$ & $12.94$ & $28.70$ & $9.00$ & $5.10$ \\
    \quad DeepSeek-R1-Distill-Llama-70B & $76.47$ & $48.24$ & $44.71$ & $67.00$ & $41.00$ & $\textbf{32.90}$ \\

    \midrule
    \textbf{\emph{Fine-tuned LLMs and VLMs:}} & &   &   &   &   &  \\
    \quad Llava1.5-13B-Uni & $68.24 \pm 18.48$ & $19.53\pm 14.98$ & $11.99\pm 10.55$ & $56.18\pm 15.68$ & $13.64\pm 11.36$ & $10.68\pm10.23$ \\
    \quad Llama2-7B-\sft{}  &  $99.76\pm 0.24$ & $65.88\pm1.05$  & $45.65 \pm 0.86$  & $99.98\pm 0.02$  & $62.64\pm 0.33$ & $53.04 \pm 0.20$ \\
    \quad Llama2-7B-\emulator{} &  $\mathbf{100}\pm 0.00$  & $69.41\pm 1.97$  & $51.53 \pm 0.44$  & $99.96\pm 0.02$  & $68.70\pm 0.49$  & $60.10\pm 0.69$ \\
    \quad Llama3-8B-\sft{} & $99.53\pm 0.29$ & $\mathbf{73.65}\pm 0.80$ & $54.12\pm 1.78$ & $99.96\pm0.04$ & $71.26\pm1.01$ & $62.72\pm1.17$\\
    \quad Llama3-8B-\emulator{} & $99.76\pm 0.24$ & $71.53 \pm 0.78$ & $\mathbf{60.23}\pm 1.01$ & $\mathbf{100}\pm 0.00$ & $\mathbf{74.92}\pm 0.60$ & $\mathbf{66.92}\pm 0.65$ \\
    \bottomrule
  \end{tabular}
  }
  \caption{Performance comparison of models on two evaluation datasets. \textbf{Bold} values indicate the highest performance in each column within the group. Fine-tuned models are trained using $5$ different random seeds and we report the mean and standard error of the performance.}
  \label{tab.main_performance}
  \vspace{-0.2cm}
\end{figure*}

\subsection{Experimental Setup} \label{sec.exp_setup}

\paragraph{Models evaluated.} We evaluate four types of models: 
(i) \emph{Base LLMs}, which include GPT-3.5~\citep{chatgpt}, GPT-4~\citep{gpt4}, Llama2~\citep{DBLP:journals/corr/abs-2307-09288}, and Llama3~\citep{llama3} family models; 
(ii) \emph{Reasoning LLMs}, which include o1~\citep{o1} and DeepSeek-R1~\citep{deepseekr1} family models; 
(iii) \emph{Base VLMs}, which include GPT-4V~\citep{gpt4}, GPT-4o~\citep{gpt4o}, Llava1.5~\citep{DBLP:journals/corr/abs-2310-03744}, InternVL2~\citep{DBLP:journals/corr/abs-2312-14238}, Qwen2-VL~\citep{DBLP:journals/corr/abs-2409-12191}, NVLM-D~\citep{DBLP:journals/corr/abs-2409-11402}, and Molmo~\citep{DBLP:journals/corr/abs-2409-17146};
(iv) \emph{Fine-tuned LLMs and VLMs}, which include fine-tuned Llava1.5, Llama2, and Llama3 models. We use ``\sft{}'' and ``\emulator{}'' as suffixes for models fine-tuned via standard supervised learning and emulator-driven methods, respectively.
Detailed versions of the evaluated models are provided in Appendix Figure~\ref{fig:model_version}. Additional evaluation and fine-tuning details are in Appendix~\ref{sec:appendix:details}.

\paragraph{Evaluation procedure and metrics.} 
We evaluate models using two datasets: \benchmarkreal{} and the synthetic dataset \benchmarksimeval{} (see Figure~\ref{tab.dataset_split}). For each task, we convert its JSON format into a natural language description using a fixed prompt template (see Figure~\ref{fig.prompt_template_nl}).
\footnote{The prompt template does not include few-shot examples or advanced prompting strategies. The evaluation of different prompting strategies is provided in Appendix~\ref{sec:appendix:prompting_strategies}.}
For multimodal models (e.g., GPT-4V, Llava1.5), we additionally provide the task image as input. Then, we prompt the model to generate Python code and extract only the code portion from the model output. Finally, we evaluate the extracted code using an emulator. We use \emph{success} as the main metric (see Section~\ref{sec.synthesis_objective}), and also consider two additional metrics: (i)~\textit{Format}, which evaluates whether the model's output adheres to the desired code format, and (ii)~\textit{No-Crash}, which evaluates whether the code runs without crashing, such as hitting walls, entering forbidden areas, or exceeding grid boundaries.

\subsection{Main Results} \label{sec.exp_results}



\begin{figure*}[h!]
    \centering
    \includegraphics[width=1\textwidth]{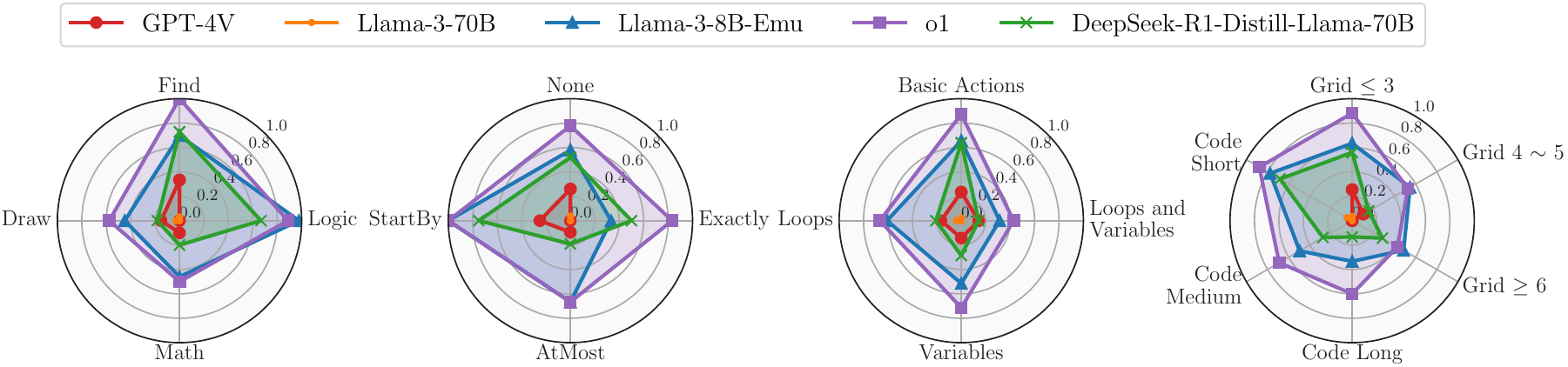}
    \begin{minipage}{0.23\textwidth}
        \begin{subfigure}[b]{1\textwidth}
            \centering
            \caption{Task Type}
            \label{fig.radar_real_performance.goal_type}
        \end{subfigure}
    \end{minipage}
    \begin{minipage}{0.25\textwidth}
        \begin{subfigure}[b]{1\textwidth}
            \centering
            \caption{Code Constraints}
            \label{fig.radar_real_performance.constraints}
        \end{subfigure}
    \end{minipage}
    \begin{minipage}{0.25\textwidth}
        \begin{subfigure}[b]{1\textwidth}
            \centering
            \caption{Code Concepts}
            \label{fig.radar_real_performance.programming_concepts}
        \end{subfigure}
    \end{minipage}
    \begin{minipage}{0.25\textwidth}
        \begin{subfigure}[b]{1\textwidth}
            \centering
            \caption{Code Length \& Grid Size}
            \label{fig.radar_real_performance.grid_size}
        \end{subfigure}
    \end{minipage}
    \caption{Comparative analysis of models' capabilities across different dimensions on \benchmarkreal{}. Each chart highlights the models' capabilities in different aspects within a dimension. Note that code length and grid size are combined in the same chart, as both indicate the difficulty levels of the tasks.}
    \label{fig.radar_real_performance}
    \vspace{-0.2cm}
\end{figure*}

As shown in Figure~\ref{tab.main_performance}, most of the evaluated models struggle significantly on our benchmark. The best performance of base models on \benchmarkreal{} and \benchmarksimeval{} is $76.47\%$ and $32.90\%$, respectively.

\paragraph{Vision capabilities provide limited benefits.}
Vision-enabled models, such as GPT-4V, show modest improvements over their text-only counterparts (GPT-4V's $20\%$ vs. GPT-4's $12.94\%$ success rate on \benchmarkreal{}), yet other VLMs continue to struggle significantly. This suggests that while vision capabilities offer some benefits, they are not the determining factor in solving our benchmark tasks. This is likely because, in our setting, the textual input can already sufficiently capture all necessary visual information required for solving the tasks. Therefore, although vision capabilities offer some incremental advantages, their overall impact on performance in our benchmark remains limited.

\paragraph{Reasoning capabilities are crucial.}
Reasoning models outperform all non-reasoning base models, with o1 achieving a $76.47\%$ success rate on \benchmarkreal{}, compared to $22.35\%$ for the best non-reasoning model.\footnote{In our evaluation, o1 failed to generate responses for $25$ tasks in \benchmarksimeval{}, so we report its performance on the remaining $975$ tasks.}
This large gap suggests that reasoning is essential for solving visual programming tasks.

\paragraph{Fine-tuning improves performance.}
Fine-tuning significantly improves model performance, especially for Llama models. Llama3-8B improves from $0\%$ to $54.12\%$ with standard fine-tuning (\sft{}), and further to $60.23\%$ with emulator-driven resampling (\emulator{}).

\subsection{Comparative Analysis Across Dimensions} \label{sec.exp_ablations}

We evaluate model performance across various dimensions to identify strengths and weaknesses. We automatically categorize each task-code pair along different dimensions (e.g., task type) and assess model capability in specific aspects (e.g., \counting{} in the task type dimension) by calculating success rates for all relevant tasks. Figure~\ref{fig.radar_real_performance} presents a comparative analysis of five representative models across distinct dimensions on \benchmarkreal{}. 
Overall, o1 and Llama3-8B-\emulator{} consistently outperform other models across almost all dimensions.
However, o1 and Llama3-8B-\emulator{} particularly struggle with draw tasks and complex code concepts (e.g., Loops and Variables), with performance degrading significantly as task complexity increases through larger grid sizes and longer code lengths.

\subsection{Failure Analysis} \label{sec.failure_analysis}

In this section, we perform failure analysis to understand the limitations of different models. 
We analyze model failures through two types of failure analysis: (i) \emph{explanation-based failure analysis}, which examines model-generated explanations to identify failure reasons, and (ii) \emph{perturbation-based failure analysis}, which evaluates performance on simplified, perturbed tasks.


\begin{figure*}[ht!]
    \centering
    \scalebox{0.695}{
    \begin{tabular}{l|rrrrrrrr|r}
    \toprule
     & Repetition & Format & Goal & \makecell[rb]{Code \\ Constraints} & \makecell[rb]{Grid \\ Constraints} & \makecell[rb]{Spatial \\ Reasoning} &  \makecell[rb]{Recursive \\ Reasoning} &  Hallucination & Success \\
    \midrule
    GPT-4V & $0.00$ & $3.53$ & $11.76$ & $7.06$ & $11.76$ & $\textbf{42.35}$  & $0.00$  & $3.53$ &  $20.00$ \\
    Llama3-70B & $34.12$ & $1.18$ & $5.88$ & $3.53$ & $8.24$ & $\textbf{44.71}$ & $0.00$ & $0.00$ & $2.35$ \\
    DeepSeek-R1-Distill-Llama-70B & $0.00$ & $0.00$ & $1.18$ & $1.18$ & $3.53$ & $23.53$ & $\textbf{25.89}$ & $0.00$ & $44.71$ \\ 
    \bottomrule
    \end{tabular}
    }
    \caption{Percentage (\%) of different failure types by analyzing model outputs on the \benchmarkreal{} dataset. \textbf{Bold} values highlight the most common failure type for each model. See Appendix~\ref{sec:appendix:failure_types} for detailed definitions of failure types.}
    \label{fig.failure_explanation}
\end{figure*}

\begin{figure*}[ht!]
    \centering
    \scalebox{1}{
    \begin{tabular}{l|c|rrr|rrr|r}
        \toprule
        &  \task{} &  $\task{}_{\codecons{}}$  & $\task{}_ {\gridcons{}}$  & $\task{}_{\spatial{}}$  & $\task{}_{\codecons{}, \gridcons{}}$  & $\task{}_{\gridcons{}, \spatial{}}$  & $\task{}_{\codecons{}, \spatial{}}$  & $\task{}_{\codecons{}, \gridcons{}, \spatial{}}$  \\
        \midrule
        GPT-4V & $0$ & $30.0$ & $30.0$ & $\textbf{50.0}$ & $50.0$ & $50.0$ & $\textbf{60.0}$ & $\textbf{60.0}$  \\
        Llama3-70B & $0$ & $0.0$ & $0.0$ & $0.0$ & $0.0$ & $0.0$ & $0.0$ & $0.0$  \\
        Llama3-8B-\sft{} & $0$ & $0.0$ & $\textbf{10.0}$ & $0.0$ & $\textbf{20.0}$ & $\textbf{20.0}$ & $0.0$ & $\textbf{30.0}$ \\
        \bottomrule
    \end{tabular}
    }
    \caption{Success rates (\%) across $80$ perturbed tasks (10 tasks $\times$ 8 perturbations). Perturbations are grouped by the number of components removed.
    The perturbations include removing code constraints ($\task{}_{\codecons{}}$), removing grid constraints ($\task{}_{\gridcons{}}$), simplifying spatial relationships ($\task{}_{\spatial{}}$), and combinations of these perturbations ($\task{}_{\codecons{}, \gridcons{}}$, $\task{}_{\gridcons{}, \spatial{}}$, $\task{}_{\codecons{}, \spatial{}}$, and $\task{}_{\codecons{}, \gridcons{}, \spatial{}}$).
    \textbf{Bold} values indicate the highest success rate for each model within each perturbation group.
    }
    \label{fig.fail_perturbation}
    \vspace{-0.2cm}
\end{figure*}

\paragraph{Explanation-based failure analysis.}
We present a failure analysis by examining output codes and explanations from GPT-4V, Llama3-70B, and DeepSeek-R1-Distill-Llama-70B. Fine-tuned models are excluded from this analysis as they generate code without explanations.
To conduct the failure analysis, we first identify common failure types. Then, we systematically analyze the explanations and output codes and manually annotate the most significant reason for each failure. In cases where multiple failure reasons are identified, we attribute the failure to the most significant reason.
The analysis results are shown in Figure~\ref{fig.failure_explanation}. Our findings show that GPT-4V and Llama3-70B struggle most with spatial reasoning, which is caused by misunderstandings of coordinates or directions after movements or turns. DeepSeek-R1-Distill-Llama-70B often fails due to recursive reasoning, where excessive reasoning produces lengthy outputs without arriving at a final answer.

\paragraph{Perturbation-based failure analysis.}
We provide failure analysis by perturbing tasks. We consider GPT-4V, Llama3-70B, and Llama3-8B-\sft{}.
We first select 10 tasks from the \benchmarkreal{} dataset that the three models consistently fail to solve.
For each task $\task{}$, we consider perturbations including removing code constraints ($\task{}_{\codecons{}}$), removing grid constraints ($\task{}_{\gridcons{}}$), simplifying spatial relationships ($\task{}_{\spatial{}}$), and combinations of these perturbations ($\task{}_{\codecons{}, \gridcons{}}$, $\task{}_{\gridcons{}, \spatial{}}$, $\task{}_{\codecons{}, \spatial{}}$, and $\task{}_{\codecons{}, \gridcons{}, \spatial{}}$). Tasks lacking certain components remain unchanged. 
As shown in Figure~\ref{fig.fail_perturbation}, GPT-4V struggles with spatial relationships. Simplifying these increases its success rate from $0\%$ to $50\%$ (see columns $\task{}$ and $\task{}_{\spatial{}}$). Conversely, Llama3-8B-\sft{} struggles with grid constraints. Removing these boosts its success rate to $10\%$ (see column $\task{}_{\gridcons{}}$), while removing code constraints and spatial relationships has no effect.
\footnote{Interestingly, Llama3-8B-\sft{} performs worse than GPT-4V in our failure analysis. This is likely because we selected tasks that all models initially failed on, which already indicates Llama3-8B-\sft{}'s limitations with these examples. When perturbed, these tasks further diverge from the training distribution. GPT-4V, with its stronger generalization capabilities, remains robust to these distribution shifts and performs better.}



\section{Concluding Discussions}
\label{sec.conclusion}

In this paper, we introduced \benchmark{}, a visual programming benchmark designed to evaluate the program synthesis capabilities of large models on visual programming tasks using the \platformAll{} environment. We found that large models struggle with visual programming tasks that require a combination of skills, even though our benchmark tasks require only basic programming skills. To improve performance, we developed a fine-tuning pipeline that involves synthetic dataset generation followed by supervised fine-tuning. This pipeline enabled the Llama3-8B model to achieve a success rate of $54.12\%$ on the benchmark tasks. Additionally, we demonstrated that leveraging emulator-driven feedback can further enhance standard fine-tuning performance by approximately $6\%$ in both the Llama3-8B and Llama2-7B models. 
Our failure analysis revealed that different models struggle with distinct issues: DeepSeek-R1-Distill-Llama-70B struggles with recursive reasoning, GPT-4V and Llama3-70B with spatial reasoning, and the fine-tuned Llama3-8B-\sft{} with grid constraints.



\section*{Limitations}

We discuss some limitations of our work and propose ideas for addressing them in the future. 
First, our benchmark focuses on basic programming skills, and future work could extend it to include more complex programming tasks. This could involve tasks that require more advanced programming concepts, such as conditionals and functions. 
Second, our emulator-driven fine-tuning provides the model with only binary feedback on the correctness of the predicted code. In the future, it would be interesting to provide more detailed feedback, such as identifying specific errors in the generated code and then using this more informative feedback to guide the fine-tuning process. 

\section*{Acknowledgments}
Funded/Co-funded by the European Union (ERC, TOPS, 101039090). Views and opinions expressed are however those of the author(s) only and do not necessarily reflect those of the European Union or the European Research Council. Neither the European Union nor the granting authority can be held responsible for them.
    \bibliography{main}
    \clearpage
    \clearpage
\appendix
{
    \allowdisplaybreaks


\section{More Details About the Datasets} \label{sec:appendix:dataset}


\begin{figure*}[h!]
    \centering
    \begin{subfigure}{0.47\textwidth}
      \includegraphics[width=\linewidth]{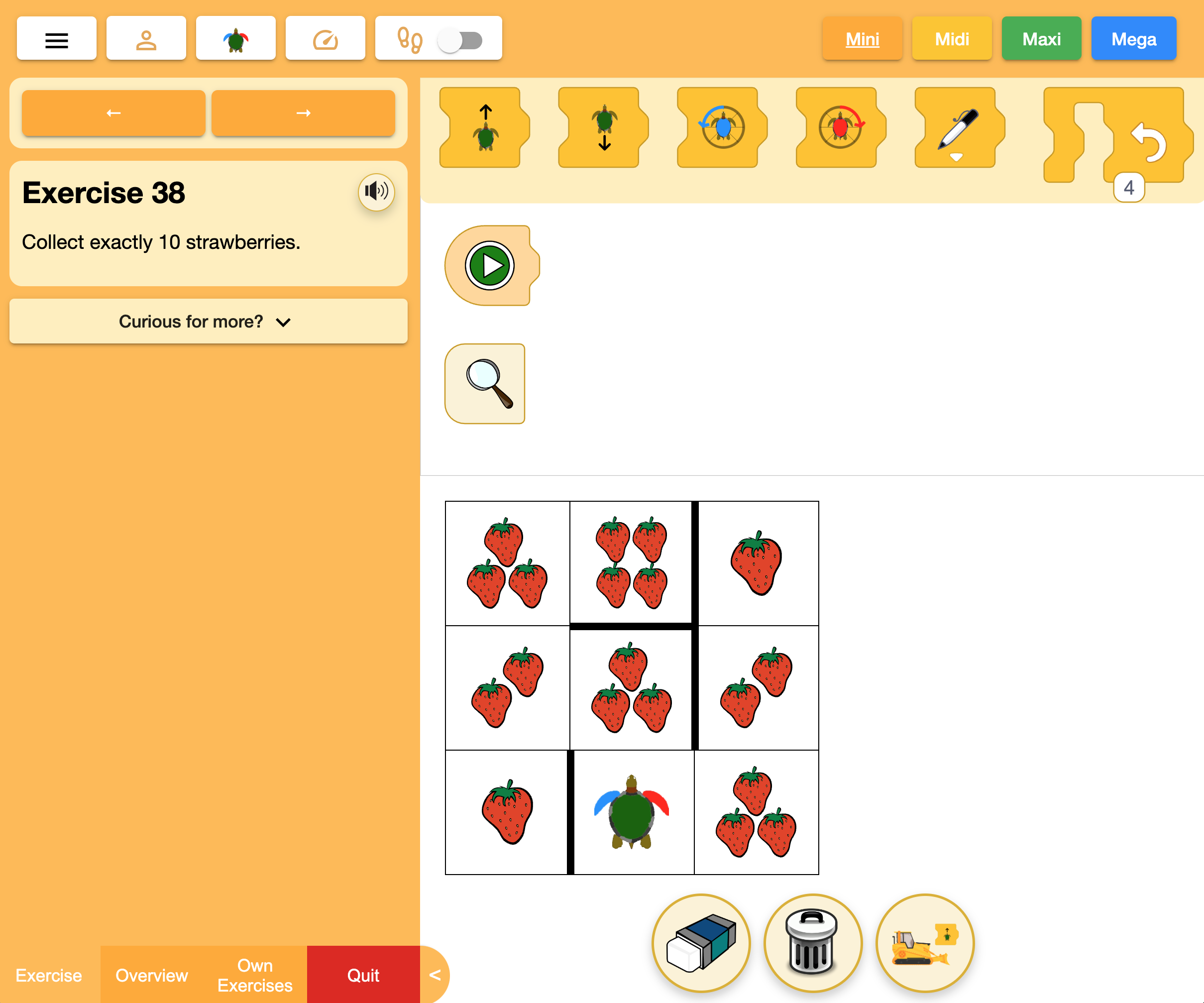}
      \caption{Task 38}
      \label{fig.pie_task_type1}
    \end{subfigure}
    \hspace{0.05\textwidth}
    \begin{subfigure}{0.46\textwidth}
      \includegraphics[width=\linewidth]{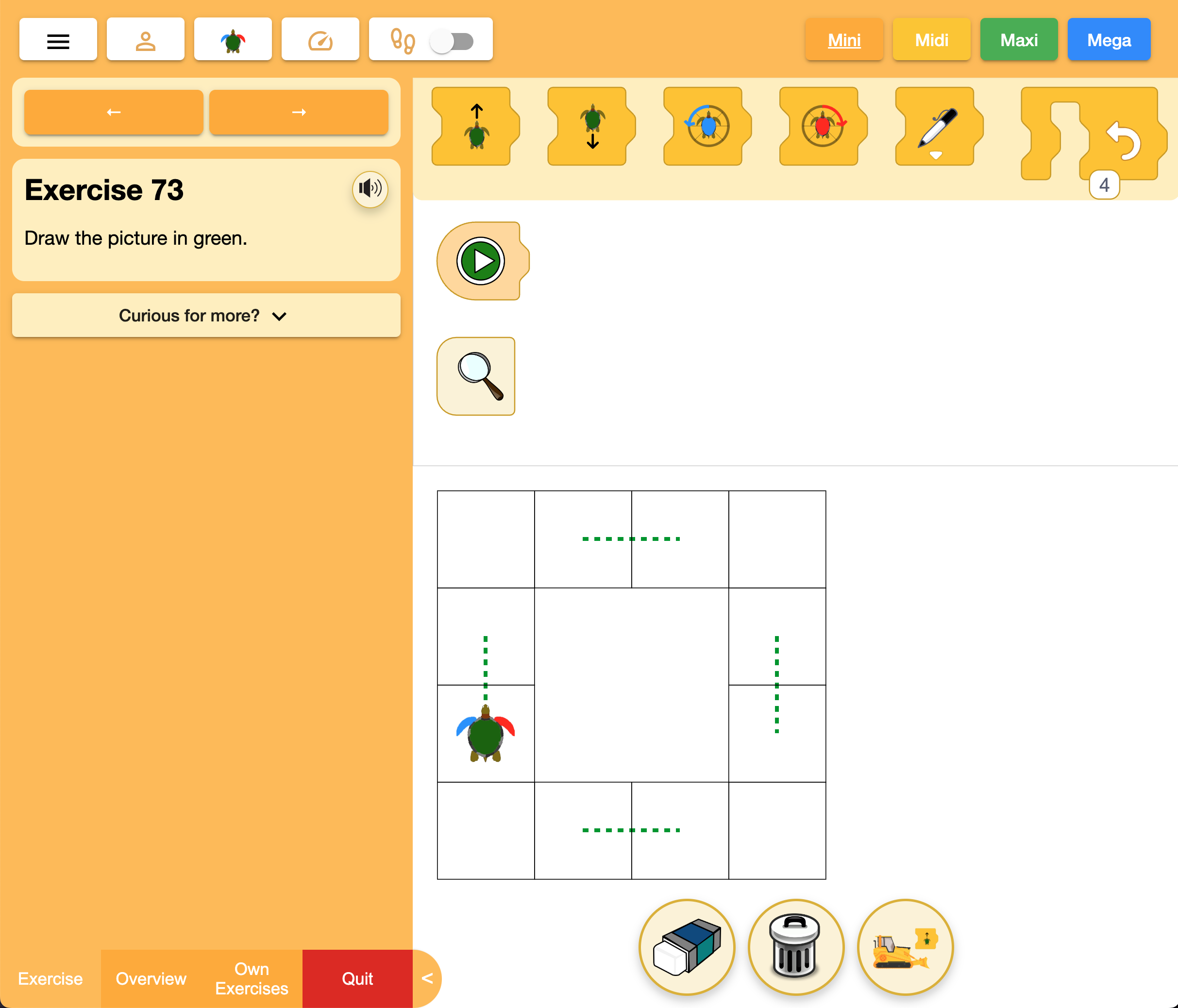}
      \caption{Task 73}
      \label{fig.pie_task_type2}
    \end{subfigure}
    \caption{Example tasks from the XLogoOnline platform. Students need to drag and drop different blocks to solve the tasks.}
    \label{fig.screenshot}
  \end{figure*}

We provide more details about the \benchmarkreal{} dataset and the synthetic \benchmarksim{} dataset.

\subsection{Details of the \benchmarkreal{} Dataset} \label{sec:appendix:real_dataset}

The real-world visual programming tasks in the \benchmarkreal{} dataset are curated from the Mini level of the XLogoOnline platform. These real-world programming tasks can be accessed and viewed at \url{https://xlogo.inf.ethz.ch/}. Figure~\ref{fig.screenshot} shows screenshots of the platform.

\subsection{Details of the Synthetic Dataset Generation} \label{sec:appendix:synthetic_dataset_generation}

In this section, we provide more details about the generation process of the synthetic dataset \benchmarksim{}. 

We use the adapted task synthesis technique~\citep{DBLP:conf/nips/AhmedCEFGRS20,chao2024xlogo} to generate a synthetic dataset. The key idea is to take a reference task and its solution code as input, and then apply symbolic execution and constraint satisfaction techniques to systematically enumerate all possible task-code outputs. The details are described as follows.

First, we manually craft a solution code for each of the $N=85$ tasks in the \benchmarkreal{} dataset, resulting in a set $\{(\task_i, \code_i)\}_{i=1}^{N}$. However, our objective is to generate a large and diverse set of tasks to train large models. To achieve this, we specify an additional parameter, the difficulty level $\diff$. This parameter enables us to generate tasks with varying levels of difficulty by specifying the desired code length, number of code constraints, and goals relative to the reference input, thereby enhancing the diversity of the dataset. The parameters are detailed as follows: 
\begin{itemize}
  \item \easy{}: The code length and number of code constraints remain the same as in the reference code and code constraints, and the goal remains unchanged. 
  \item \medium{}: The code length is increased by 1 or 2 additional commands compared to the reference code, while the number of code constraints and the goal remain the same as in the reference task $\task$. 
  \item \hard{}: The code length is increased by up to 2 additional commands, one more code constraint is added compared to the reference code constraints, and the goal may be modified.
\end{itemize}
Note that the difficulty levels mentioned above indicate the relative difficulty of the generated tasks compared to the reference task, not the absolute difficulty of the tasks.

Given the reference input $(\task, \code, \diff)$, we begin by enumerating all possible codes, code constraints, and goals that meet the specified difficulty levels. To achieve this, we first create templates for the code, constraints, and goals, each containing placeholders. These placeholders are then populated with specific values using an SMT-based constraint solver~\citep{DBLP:conf/tacas/MouraB08}. This process allows us to generate all possible combinations of code, constraints, and goals that align with the desired difficulty levels. 

Next, we generate task-code pairs by combining the previously generated code, code constraints, and goals with corresponding grid worlds. To generate these grid worlds, we symbolically execute the previously generated code within an empty grid, constructing elements like walls and target items to ensure the grid can be successfully solved by the code. After the grid world is constructed, it is combined with the corresponding code, code constraints, and goal to form a task-code pair.

In implementation, we generate up to 3,000 tasks for each combination of code, code constraints, and goals. Subsequently, we sample 500 tasks from the pool of all generated tasks for each $(\task, \code, \diff)$, resulting in up to $500\ \text{tasks} \times 3\ \text{difficulty levels} = 1,500$ tasks for each reference input $(\task, \code)$. This process is repeated for all reference inputs in the dataset, resulting in a total of up to $85 \times 1,500 = 127,500$ tasks. Finally, we apply the processing steps described in the main paper to generate the synthetic dataset, resulting in the final dataset, \benchmarksim{}, containing $89,053$ tasks and solution codes. 

To run the adapted task synthesis technique, we use a 12-core, 3 GHz Intel Xeon E7-8857 CPU, with parallelization across 8 cores under a 64-bit Debian operating system.

\begin{figure*}[ht!]
  \centering
  \begin{minipage}[t]{0.185\linewidth}
    \centering
    \begin{subfigure}[b]{\linewidth}
      \caption*{\scriptsize Collect all red shapes without standing on the green.}
      \includegraphics[width=\textwidth]{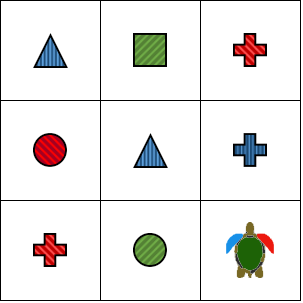}
      \vspace{-4pt}
    \end{subfigure}
    \begin{subfigure}[b]{\linewidth}
      \textbf{\scriptsize Solution Code:}
      \begin{lstlisting}
def Run():
  $\fd{}$
  $\fd{}$
  $\bk{}$
  $\lt{}$
  $\fd{}$
  $\fd{}$
  $\lt{}$
  $\fd{}$
  $$
  $$
  $$
      \end{lstlisting} 
    \end{subfigure}
  \end{minipage}
  \hspace{0.005\linewidth}
  \begin{minipage}[t]{0.185\linewidth}
    \centering
    \begin{subfigure}[b]{\linewidth}
      \caption*{\scriptsize Collect exactly 5 strawberries.\\}
      \includegraphics[width=1\textwidth]{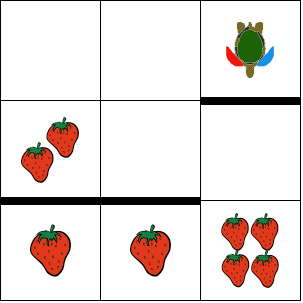}
      \vspace{-4pt}
    \end{subfigure}
    \begin{subfigure}[b]{\linewidth}
      \textbf{\scriptsize Solution Code:}
      \begin{lstlisting}
def Run():
  $\rt{}$
  $\fd{}$
  $\lt{}$
  $\fd{}$
  $\lt{}$
  $\fd{}$
  $\rt{}$
  $\fd{}$
  $\rt{}$
  $\fd{}$
  $$
      \end{lstlisting} 
    \end{subfigure}
  \end{minipage}
  \hspace{0.005\linewidth}
  \begin{minipage}[t]{0.185\linewidth}
    \centering
    \begin{subfigure}[b]{\linewidth}
      \centering
      \caption*{\scriptsize Draw the picture in yellow. Use at most 8 commands.}
      \includegraphics[width=0.97\textwidth]{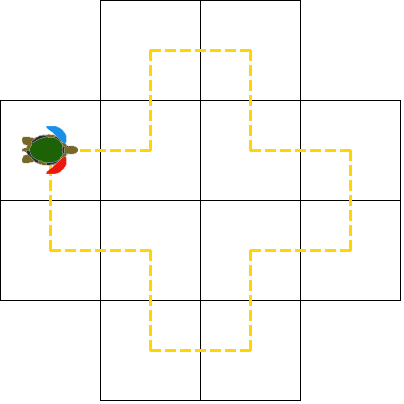}
      \vspace{12pt}
    \end{subfigure}
        \begin{subfigure}[b]{\linewidth}
          \textbf{\scriptsize Solution Code:}
          \begin{lstlisting}
def Run():
  $\setpc{}$("yellow")
  $\Repeat{}$ i in range(4):
    $\fd{}$
    $\lt{}$
    $\fd{}$
    $\lt{}$
    $\bk{}$
    $\lt{}$
    $$
    $$
    $$

          \end{lstlisting} 
    \end{subfigure}
  \end{minipage}
  \hspace{0.005\linewidth}
  \begin{minipage}[t]{0.185\linewidth}
    \centering
    \begin{subfigure}[b]{\linewidth}
      \caption*{\scriptsize Draw the picture. Use at most 8 commands.}
      \includegraphics[width=\textwidth]{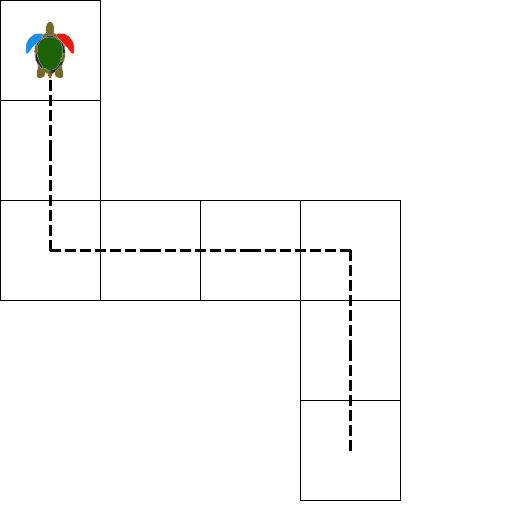}
      \vspace{-4pt}
    \end{subfigure}
        \begin{subfigure}[b]{\linewidth}
          \textbf{\scriptsize Solution Code:}
          \begin{lstlisting}
def Run():
  $\Repeat{}$ i in range(2):
    $\bk{}$
  $\rt{}$
  $\Repeat{}$ i in range(3):
    $\fd{}$
  $\rt{}$
  $\Repeat{}$ i in range(2):
    $\fd{}$
    $$
    $$
    $$
          \end{lstlisting} 
        \end{subfigure}
  \end{minipage}
  \hspace{0.005\linewidth}
  \begin{minipage}[t]{0.185\linewidth}
    \centering
    \begin{subfigure}[b]{\linewidth}
      \caption*{\scriptsize Find the strawberry. Use at most 6 commands.}
      \includegraphics[width=1\textwidth]{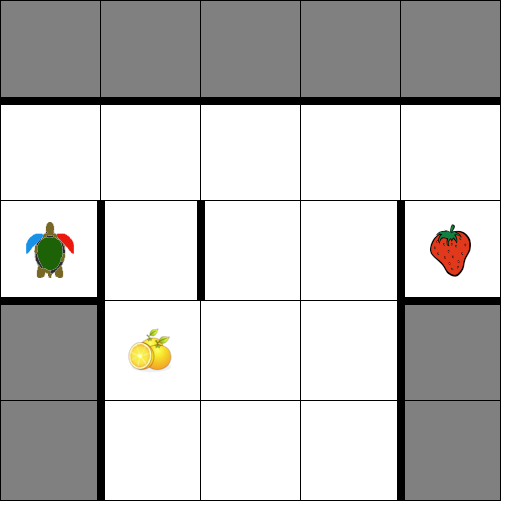}
      \vspace{-4pt}
    \end{subfigure}
        \begin{subfigure}[b]{\linewidth}
          \textbf{\scriptsize Solution Code:}
          \begin{lstlisting}
def Run():
  $\fd{}$
  $\rt{}$
  $\Repeat{}$ i in range(4):
    $\fd{}$
  $\rt{}$
  $\fd{}$
  $$
  $$
  $$
  $$
  $$
          \end{lstlisting} 
        \end{subfigure}
  \end{minipage}
  \caption{Examples of synthetic tasks and their corresponding solution codes in \benchmarksimeval{}. Note that while the synthesized solution codes are correct, they may not use the minimum number of commands.}
  \label{fig.example_tasks_simeval}
\end{figure*}

\subsection{Quality of the Datasets}

The quality of the datasets is crucial for the success of the models trained on them. Therefore, we provide more details about the quality of the datasets. We mainly use the following two datasets for evaluation: 
\begin{enumerate}[leftmargin=*]
    \item \benchmarkreal{} dataset (85 samples): This dataset was derived from the visual programming platform \platformAll{}. The tasks included in this platform were meticulously crafted by experts and have been used by tens of thousands of students every year~\citep{DBLP:conf/issep/HromkovicSS17,DBLP:journals/eatcs/Staub21}. Given this extensive use and expert involvement, the quality of the tasks in this dataset is guaranteed.
    
    \item \benchmarksimeval{} dataset (1000 samples): This dataset was synthetically generated. However, we ensure data quality by implementing the following checks: (i) we have removed any duplicate task-code pairs; (ii) we have conducted a correctness check on the generated solution codes using the emulator; and (iii) we have excluded any task-code pairs present in the \benchmarkreal{} dataset from this synthetic dataset. In Figure~\ref{fig.example_tasks_simeval}, we show examples of the tasks in this dataset.
\end{enumerate}


\begin{figure*}[h]
    \centering
    \begin{tabular}{lcccc}
    \toprule
    & Visual Appeal & Grid Elements Utility & Code Quality & Overall Quality \\
    \midrule
    \benchmarkreal{} & $1.00$ & $1.00$ & $1.00$ & $1.00$ \\
    \benchmarksimeval{}  & $0.97$ & $0.94$ & $0.89$ & $0.84$ \\
    \bottomrule
    \end{tabular}
    \caption{Quality annotation results for \benchmarkreal{} and \benchmarksimeval{} datasets. For \benchmarkreal{}, we annotate all 85 samples, while for \benchmarksimeval{}, we randomly sample 5\% of the dataset for annotation.}
    \label{fig.quality}
  \end{figure*}

To further demonstrate the quality of our datasets, we conduct a quality annotation for both datasets. Specifically, we annotate the quality of all 85 samples in the \benchmarkreal{} dataset and randomly sample 5\% of tasks from the \benchmarksimeval{} dataset for annotation. The following rubrics are used to evaluate each (task, code) pair:
\begin{enumerate}[leftmargin=*]
    \item \emph{Visual appeal}
    \begin{itemize}
        \item 0: Poor - The visual grid is highly unappealing.
        \item 0.5: Acceptable - The visual grid is moderately appealing.
        \item 1: Excellent - The visual grid is highly appealing.
    \end{itemize}
    
    \item \emph{Grid elements utility}
    \begin{itemize}
        \item 0: Poor - The distractors are neither useful nor reasonably positioned.
        \item 0.5: Acceptable - Some distractors are useful, while others lack utility.
        \item 1: Excellent - Most, if not all, distractors are strategically useful and sensibly placed.
    \end{itemize}
    
    \item \emph{Code quality}
    \begin{itemize}
        \item 0: Poor - The code is of poor quality, unable to solve the task, or violates code constraints.
        \item 0.5: Acceptable - The code can solve the task but contains some unnecessary commands.
        \item 1: Excellent - The code solves the task, meets code constraints, and has no redundant commands.
    \end{itemize}
    
    \item \emph{Overall quality}: Calculated as the minimum score across visual appeal, grid elements utility, and code quality.
\end{enumerate}

The results in Figure~\ref{fig.quality} demonstrate that the overall quality of the \benchmarkreal{} dataset is excellent. The \benchmarksimeval{} dataset, with an overall quality score of $0.84$, exceeds the acceptable threshold (score = $0.5$) and approaches the level of excellence (score = $1.0$). Additionally, during the quality annotation, we do not find any (task, code) pair where the task is unsolvable or the code fails to successfully solve the task.
 
    \clearpage


\section{More Details of the Failure Analysis, Fine-tuning, and Evaluation} \label{sec:appendix:details}

In this section, we provide more details about the failure analysis, fine-tuning, and evaluation.

\subsection{Details of Failure Analysis} \label{sec:appendix:failure_types}

We provide details of the explanation-based failure analysis. To conduct the explanation-based failure analysis, we first identify common failure types. In cases where multiple failure reasons are identified, we attribute the failure to the most significant cause. These failure types are defined as follows: 
\begin{itemize}[leftmargin=*]
    \item \emph{Repetition}: generating the same code sequences repeatedly; 
    \item \emph{Format}: producing code with incorrect formatting, including the use of disallowed commands; 
    \item \emph{Goal}: misinterpreting the goal or attempting to devise a tricky approach to achieve the goal; 
    \item \emph{Code constraints}: failing to adhere to specified code constraints while solving the task; 
    \item \emph{Grid constraints}: attempting to solve the task while ignoring walls, forbidden cells, or grid boundaries; 
    \item \emph{Spatial reasoning}: misunderstanding coordinates or directions after movements or turns; 
    \item \emph{Recursive reasoning}: failing to arrive at a final answer due to excessive or circular reasoning, leading to lengthy responses that exceed maximum token generation limits;
    \item \emph{Hallucination}: generating non-existent items or code commands. 
\end{itemize}

\subsection{Details of the Evaluation} \label{sec:appendix:evaluation_details}

\paragraph{Versions of evaluated models.} The versions of the evaluated models are provided in Figure~\ref{fig:model_version}.


\begin{figure*}[ht!]
    \centering
    \scalebox{0.9}{
    \begin{tabular}{ll}
      \toprule
      \textbf{Model} & \textbf{Version} \\
      \midrule
      \textbf{\emph{Base LLMs (text-only):}} & \\
      \quad GPT-3.5 & gpt-3.5-turbo-0125~\cite{chatgpt} \\
      \quad GPT-4 & gpt-4-turbo-2024-04-09~\cite{gpt4} \\
      \quad Llama2-7B & meta-llama/Llama-2-7b-chat~\cite{DBLP:journals/corr/abs-2307-09288} \\
      \quad Llama2-13B & meta-llama/Llama-2-13b-chat~\cite{DBLP:journals/corr/abs-2307-09288} \\
      \quad Llama2-70B & meta-llama/Llama-2-70b-chat~\cite{DBLP:journals/corr/abs-2307-09288} \\
      \quad Llama3-8B & meta-llama/Meta-Llama-3-8B-Instruct~\cite{llama3} \\
      \quad Llama3-70B & meta-llama/Meta-Llama-3-70B-Instruct~\cite{llama3} \\

      \midrule
      \textbf{\emph{Base VLMs (text + vision):}} & \\
      \quad GPT-4o & gpt-4o-2024-11-20~\cite{gpt4o} \\
      \quad GPT-4V & gpt-4-turbo-2024-04-09~\cite{gpt4} \\
      \quad Llava1.5-7B & liuhaotian/llava-v1.5-7b~\cite{DBLP:journals/corr/abs-2310-03744} \\
      \quad Llava1.5-13B & liuhaotian/llava-v1.5-13b~\cite{DBLP:journals/corr/abs-2310-03744} \\
      \quad InternVL2-8B & OpenGVLab/InternVL2-8B~\cite{DBLP:journals/corr/abs-2312-14238} \\
      \quad InternVL2-76B-Llama3-76B & OpenGVLab/InternVL2-Llama3-76B~\cite{DBLP:journals/corr/abs-2312-14238} \\
      \quad Qwen2-VL-7B & Qwen/Qwen2-VL-7B-Instruct~\cite{DBLP:journals/corr/abs-2409-12191} \\
      \quad Qwen2-VL-72B & Qwen/Qwen2-VL-72B-Instruct~\cite{DBLP:journals/corr/abs-2409-12191} \\
      \quad NVLM-D & nvidia/NVLM-D-72B~\cite{DBLP:journals/corr/abs-2409-11402} \\
      \quad Molmo-72B & allenai/Molmo-72B-0924~\cite{DBLP:journals/corr/abs-2409-17146} \\
      \quad Molmo-7B-D & allenai/Molmo-7B-D-0924~\cite{DBLP:journals/corr/abs-2409-17146} \\

      \midrule
      \textbf{\emph{Reasoning LLMs (text-only):}} & \\
      \quad o1 & o1-2024-12-17~\cite{o1} \\
      \quad DeepSeek-R1-Distill-Llama-8B & DeepSeek/DeepSeek-R1-Distill-Llama-8B~\cite{deepseekr1} \\
      \quad DeepSeek-R1-Distill-Llama-70B & DeepSeek/DeepSeek-R1-Distill-Llama-70B~\cite{deepseekr1} \\

      \midrule
      \textbf{\emph{Fine-tuned LLMs and VLMs:}} & \\
      \quad Llava1.5-13B-\sft{} & liuhaotian/llava-v1.5-13b (supervised fine-tuning) \\
      \quad Llama2-7B-\sft{} & meta-llama/Llama-2-7b (supervised fine-tuning) \\
      \quad Llama2-7B-\emulator{} & meta-llama/Llama-2-7b (emulator-driven fine-tuning) \\
      \quad Llama3-8B-\sft{} & meta-llama/Meta-Llama-3-8b (supervised fine-tuning) \\
      \quad Llama3-8B-\emulator{} & meta-llama/Meta-Llama-3-8b (emulator-driven fine-tuning) \\
      \bottomrule
    \end{tabular}
    }
    \caption{Evaluated models and their versions.}
    \label{fig:model_version}
\end{figure*}

\paragraph{Details of evaluation.}
All models are queried with a temperature of $0$, except for the DeepSeek-R1 family models, which are queried with a temperature of $0.6$. 
To evaluate GPT family models, we use the OpenAI API with a temperature of $0$. 
For base LLMs, base VLMs, and fine-tuned models, we use the vLLM~\citep{DBLP:conf/sosp/KwonLZ0ZY0ZS23} inference engine with $2$ A100 GPUs, using a temperature of $0$ and \texttt{max\_num\_seqs} of $2$. We find that a smaller \texttt{max\_num\_seqs} value slows down inference speed but improves performance. Therefore, we choose a \texttt{max\_num\_seqs} value of $2$ to balance performance and speed for inference. 
For the DeepSeek-R1 family models, we use the vLLM inference engine with $4$ A100 GPUs for DeepSeek-R1-Distill-Llama-8B and $8$ A100 GPUs for DeepSeek-R1-Llama-70B. We set a temperature of $0.6$ and \texttt{max\_num\_seqs} to $2$, with a maximum of $8192$ tokens to enable extra reasoning tokens. 
For the o1 model, due to high cost and budget constraints, we set the maximum token generation limit (\texttt{max\_completion\_tokens}) to $4096$ and the \texttt{reasoning\_effort} to \texttt{Medium} when querying.
After inference, we use the emulator to evaluate the models' success rates over the evaluation datasets. 


\subsection{Details of Fine-tuning} \label{sec:appendix:finetuning_details}

\paragraph{Details of fine-tuning Llama family models.}
For Llama family models, we choose non-instruction-tuned versions for fine-tuning because the base models will be fine-tuned to generate code, without requiring instruction-following capabilities. We use LoRA for parameter-efficient fine-tuning~\citep{DBLP:conf/iclr/HuSWALWWC22}. To find the best LoRA rank and scaling factor, we experimented with ranks of $8$, $16$, $32$, and $64$, using a scaling factor $\alpha$ four times the rank in each case. We found that ranks of $32$ and $64$ provide the best performance. Consequently, we use a rank of $32$ and a scaling factor of $128$ for all fine-tuning experiments. Fine-tuning is performed with a batch size of $4$ and a learning rate of $1 \times 10^{-4}$. All fine-tuning experiments are conducted on an internal cluster using $4$ A100 GPUs. Each epoch of fine-tuning for the Llama3-8B and Llama2-7B models takes approximately $3.75$ hours. In our experiments, all fine-tuned Llama models are trained for $8$ epochs, as we observed that the validation dataset loss stabilizes around epoch $8$. We train all fine-tuned Llama models using $5$ different random seeds.

\paragraph{Details of fine-tuning Llava family model.}
We perform standard supervised fine-tuning on Llava1.5-13B~\citep{DBLP:journals/corr/abs-2310-03744}. To do this, we follow the default fine-tuning setup and code provided by the authors. Specifically, we use LoRA with a rank of $128$ and a scaling factor of $256$ for fine-tuning Llava1.5-13B. During fine-tuning, we use a batch size of $16$, a learning rate of $2 \times 10^{-4}$, and a maximum sequence length of $2048$. We fine-tune the Llava model for $3$ epochs on the $87$k training dataset using $5$ different random seeds, utilizing $4$ A100 GPUs.

\paragraph{Details of emulator-driven fine-tuning.} For emulator-driven fine-tuning, we use the same hyperparameters and setup as standard fine-tuning, with the exception of resampling every $3$ epochs. Specifically, we resample the training dataset based on the emulator's evaluation results every $3$ epochs. 
To save time and resources, we start from the checkpoint of the fine-tuned models without resampling at epoch $3$. We then reuse this checkpoint to continue fine-tuning for $5$ additional epochs using emulator-driven resampling, resulting in a total of $8$ epochs. Emulator-driven resampling requires calculating a weight for each training sample, which involves inference over the entire training dataset. For inference, we use the vLLM inference engine~\citep{DBLP:conf/sosp/KwonLZ0ZY0ZS23} with \texttt{max\_num\_seqs} of $8$, batch size of $2$, and temperature of $0$. In this setting, a single iteration of inference and resampling on the $87$k training dataset takes approximately $8$ hours. After inference, we use the emulator to evaluate the correctness of the model's predicted code. Based on this evaluation, we calculate the weight for each training sample using a value of $\beta=1$.

    \clearpage


\section{Additional Experiments and Results} \label{sec:appendix:experiments}

In this section, we present additional experiments and results.

\subsection{Influence of Prompting Strategies} \label{sec:appendix:prompting_strategies}



\begin{figure}[ht!]
    \centering
    \scalebox{0.95}{
    \begin{tabular}{lrrrr}
            \toprule
             & Vanilla & 3-shot & 3-shot + CoT \\
            \midrule
            GPT-4 & $12.94$ & $10.59$ &  $18.82$ \\
            GPT-4V & $20$ & $14.12$ & $15.29$ \\
            \bottomrule
    \end{tabular}
    }
    \caption{Success rates (\%) of GPT-4 and GPT-4V with different prompting strategies on the \benchmarkreal{} dataset.}
    \label{fig.prompting}
\end{figure}

We conduct experiments on different prompting strategies to investigate their effectiveness in our benchmark. 
We consider the following prompting strategies:
(i) \emph{Vanilla} is the prompt without any additional examples or chain-of-thoughts;
(ii) \emph{3-shot} is the prompt with 3-shot examples~\citep{DBLP:conf/nips/BrownMRSKDNSSAA20};
(iii) \emph{3-shot + CoT} is the prompt with the 3-shot examples and a step-by-step chain-of-thought (CoT) for each example~\citep{DBLP:conf/nips/Wei0SBIXCLZ22}.
Note that the 3-shot examples are manually designed to ensure they cover most skills, including math, logic, drawing, basic actions, variables, loops, and code constraints. These same 3-shot examples are used to prompt all tasks for both \emph{3-shot} and \emph{3-shot + CoT} prompting.

The results are shown in Figure~\ref{fig.prompting}. We observe that \emph{3-shot} prompting by itself is not notably effective. However, when combined with CoT, it leads to performance improvements, though these gains are limited.
We speculate that this is due to the nature of our visual programming tasks, which require long-term path planning, an understanding of spatial relationships, and accurate prediction of the consequences of actions. These elements are typically absent from the training data, making it difficult for the model to leverage in-context learning to solve unfamiliar visual programming tasks.

\subsection{Influence of Task Representations}



\begin{figure*}[ht!]
    \begin{subfigure}[b]{0.45\textwidth}
      \centering
      \scalebox{0.8}{
      \begin{tabular}{lcc}
        \toprule
        & \multicolumn{2}{c}{Success Rates (\%)} \\
        \cmidrule{2-3}
        &  NL & ASCII \\
        \midrule
        \textbf{Base models} & &  \\
        \quad GPT-4 & $\mathbf{12.94}$  & $5.88$   \\
        \quad Llama3-70B & $\mathbf{2.35}$ & $1.18$ \\
        \midrule
        \textbf{Fine-tuned models} &  &  \\
        \quad Llama3-8B-\sft{} &  $\mathbf{54.12} \pm 1.78$ & $ 53.18 \pm 1.01 $ \\
        \bottomrule
      \end{tabular}
      }
      \vspace{20pt}
      \caption{Performance of base and fine-tuned models with NL and ASCII prompts.}
      \label{tab.task_representation}
    \end{subfigure}
    \hspace{0.05\textwidth}
    \begin{subfigure}[b]{0.45\textwidth}
      \centering
      \includegraphics[width=0.90\textwidth]{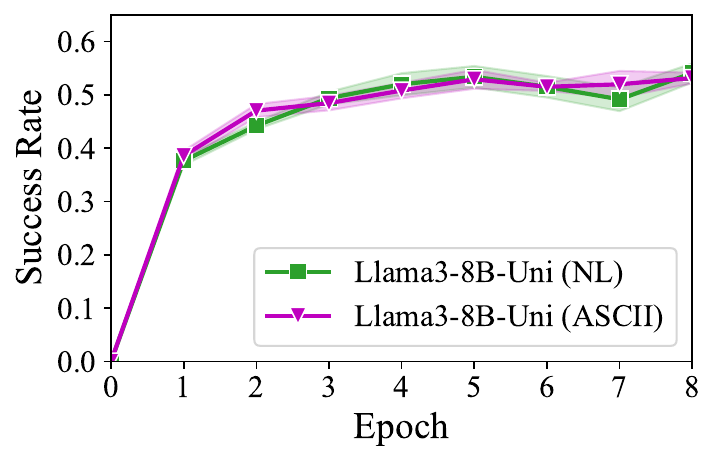}
      \caption{Performance of Llama3-8B-\sft{} across epochs with NL and ASCII prompts.}
      \label{fig.train_epoch_nl_ascii}
    \end{subfigure}
    \caption{Influence of task representations on model performance. We compare the performance of base models and fine-tuned models using natural language (NL) and ASCII prompts, respectively. (a) shows the success rates of base and fine-tuned models. (b) shows the performance of fine-tuned models across different epochs. Natural language prompts lead to better performance in base models. However, the fine-tuned Llama3-8B-Uni performs similarly with both NL and ASCII prompts.}
    \label{fig.task_representation}
\end{figure*}

In this section, we investigate the influence of natural language and ASCII representations on model performance. For visual programming tasks, the 2-dimensional grid can be represented in various ways, including natural language descriptions, ASCII-based representations, and images. For the ASCII representation, we developed a template to represent the task's visual grid using ASCII characters. These ASCII characters are then provided to the model as a replacement for the natural language descriptions of the visual grid, both for fine-tuning and evaluation. An example of an ASCII-based prompt is shown in Figure~\ref{fig.prompt_template_ascii.example1}. 

The evaluation results are shown in Figure~\ref{fig.task_representation}. Our results indicate that GPT-4 and Llama3-70B perform better with natural language (NL) representations. This might be due to their predominant training on natural language data. However, the fine-tuned Llama3-8B-\sft{} model performs similarly with both NL and ASCII prompts, with final success rates of $54.12\%$ and $53.18\%$, respectively. 

In Figure~\ref{fig.train_epoch_nl_ascii}, we show Llama3-8B-\sft{}'s performance across different epochs with NL and ASCII prompts. We find that the performance of Llama3-8B-\sft{} with NL and ASCII prompts converges at a similar rate, suggesting that fine-tuning helps the model adapt to ASCII-based task representations, making task representations less critical for fine-tuned models in our visual programming domain.


\subsection{Fine-tuning Performance Across Different Epochs}

Figure~\ref{fig.train_epoch_real} illustrates the performance of fine-tuned models across different epochs. For the emulator-driven fine-tuning (\emulator{}), we adjust the resampling interval to every three epochs, specifically at epochs 3 and 6. At epoch 3, we reuse the checkpoint from the standard fine-tuning (\sft{}) to save time and resources. As a result, the performance of the emulator-driven fine-tuning (\emulator{}) matches that of the corresponding standard fine-tuning (\sft{}) up until epoch 3. Then, an emulator-driven resampling is performed at epoch 3, leading to further performance improvements compared to models without resampling. 
Notably, at the end of training, Llama2-7B-\emulator{} achieves performance close to that of Llama3-8B-\sft{}, despite the latter being fine-tuned on a more advanced base model. This demonstrates the effectiveness of the curriculum designed by emulator-driven resampling in enhancing the performance of standard fine-tuning.



\begin{figure*}[t!]
  \begin{subfigure}[b]{0.45\textwidth}
    \centering
    \includegraphics[width=0.8\textwidth]{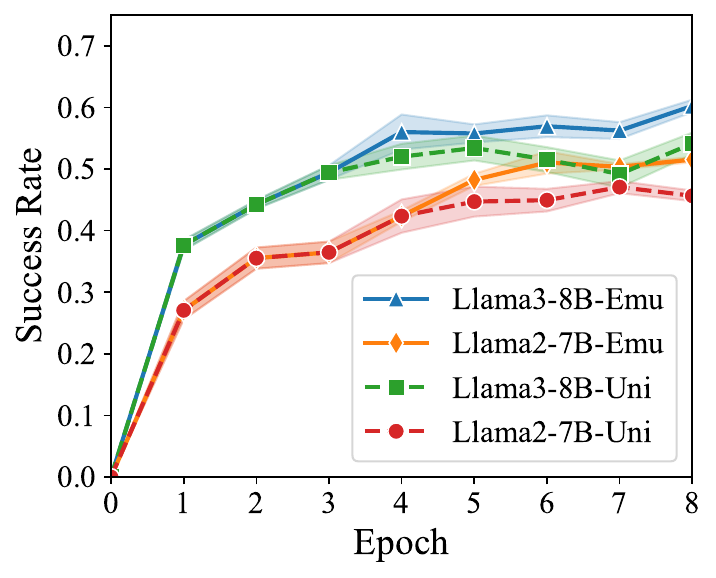}
    \caption{\benchmarkreal{} dataset.}
    \label{fig.train_epoch_real}
  \end{subfigure}
  \hspace{0.05\textwidth}
  \begin{subfigure}[b]{0.45\textwidth}
    \centering
    \includegraphics[width=0.8\textwidth]{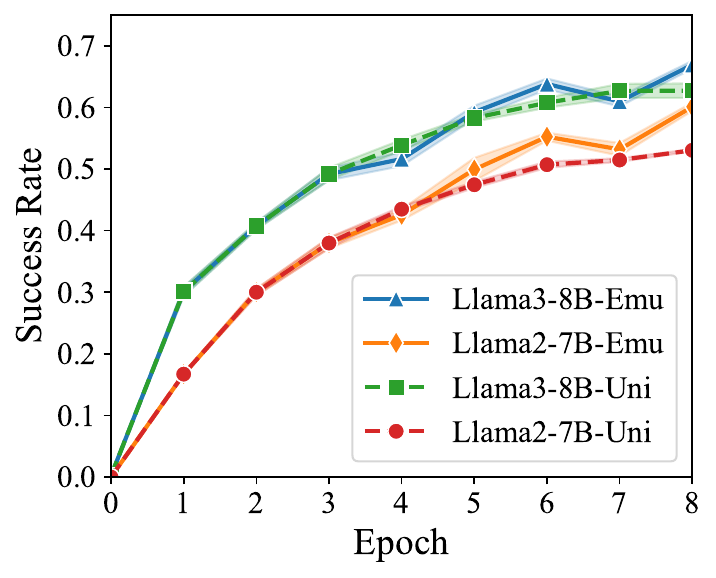}
    \caption{\benchmarksimeval{} dataset.}
    \label{fig.train_epoch_syn}
  \end{subfigure}
  \caption{Fine-tuning performance across different epochs on two evaluation datasets. (a) shows the performance of fine-tuned models across different epochs on the evaluation dataset \benchmarkreal{}. (b) shows the fine-tuning performance across different epochs on the synthetic evaluation dataset \benchmarksimeval{}.}
\end{figure*}

In Figure~\ref{fig.train_epoch_syn}, we show the fine-tuning performance across different epochs on the synthetic evaluation dataset \benchmarksimeval{}. This synthetic evaluation dataset exhibits the same distribution as the training dataset due to our splitting method. Emulator-driven resampling is performed at epochs 3 and 6 for both Llama3-8B-\emulator{} and Llama2-7B-\emulator{}. We find that standard fine-tuning without resampling leads to a smooth increase in performance across epochs, as seen in the Llama3-8B-\sft{} and Llama2-7B-\sft{} curves. In contrast, emulator-driven fine-tuning with resampling shows slight performance fluctuations, particularly in the epochs immediately following resampling (i.e., epochs 4 and 7). The fluctuations in emulator-driven fine-tuning might be due to the resampling process altering the distribution of the training data, leading to a temporary drop in performance. However, in later epochs after resampling (e.g., epoch 8), the performance of resampling-based models outperforms that of the standard fine-tuning models, indicating the effectiveness of emulator-driven fine-tuning in improving fine-tuning performance.


\begin{figure*}[ht!]
    \centering
    \scalebox{1}{
    \begin{tabular}{lrr}
    \toprule
     & {\benchmarkreal{} ($10$ tasks)} & {\small \benchmarksimeval{} ($176$ tasks)} \\ 
    \midrule
    Llama3-70B & $0.00$ & $0.00$ \\ 
    Llama3-8B-\sft{} (no-math) & $10.00\pm 10.00$ & $6.25 \pm 1.18$ \\ 
    Llama3-8B-\sft{} & $\textbf{40.00} \pm 5.48$ & $\textbf{38.98} \pm 1.82$ \\ 
    \bottomrule
    \end{tabular}
    }
    \caption{Success rates (\%) of models on math tasks. Success rates of fine-tuned models are reported as mean and standard error across five seeds.}
    \label{fig.ood}
\end{figure*}


\begin{figure*}[h!]
    \centering
    \scalebox{1}{
    \begin{tabular}{lcccc}
    \toprule
    & HumanEval & HumanEval+ & MBPP & MBPP+ \\
    \midrule
    Llama3-8B (Base) & $36.6\%$ & $31.1\%$ & $62.4\%$ & $52.6\%$ \\
    Llama3-8B-Uni (Fine-tuned) & $33.5\%$ & $26.8\%$ & $57.9\%$ & $46.8\%$ \\
    \midrule
    $\Delta$ (Fine-tuned - Base) & $-3.1\%$ & $-4.3\%$ & $-4.5\%$ & $-5.8\%$ \\
    \bottomrule
    \end{tabular}
    }
    \caption{Pass@1 performance of Llama3-8B (Base) and the Llama3-8B-\sft{} (fine-tuned) on other program synthesis benchmarks, including HumanEval, HumanEval+, MBPP, and MBPP+. Fine-tuning on the \benchmarksim{} dataset leads to a performance drop of $3 \sim 6\%$ on these program synthesis benchmarks.}
    \label{fig.ood_performance_program_synthesis_benchmarks}
\end{figure*}

\subsection{Can Fine-tuned Models Learn Transferable Skills?} \label{sec.ood}
We explore whether fine-tuned models can develop transferable skills to solve tasks that are not seen during training. To investigate this, we first exclude all tasks involving math skills (e.g., Task 38 in Figure~\ref{fig.example_tasks}) from the training dataset, resulting in a reduced training dataset with 72k samples. Then, we fine-tune Llama3-8B on this reduced dataset using standard supervised learning, referring to the resulting model as \emph{Llama3-8B-\sft{} (no-math)}. 
Next, we evaluate this model exclusively on math tasks from the evaluation datasets. The results are shown in Figure~\ref{fig.ood}. Our results reveal that Llama3-8B-\sft{} (no-math) outperforms Llama3-70B, despite neither model being trained on math tasks. This suggests that the fine-tuned Llama3-8B-Uni (no-math) acquires certain transferable skills. However, compared to Llama3-8B-\sft{}, which was trained on the full dataset including math tasks, the no-math version performs much worse. This indicates that while Llama3-8B-\sft{} (no-math) learns some generalizable skills, it is less effective than the model trained directly on data that includes those skills.

\subsection{Impact of Domain-Specific Fine-Tuning on Other Benchmarks} \label{sec:appendix:fine_tuning_impact_other_benchmarks}

We have shown that fine-tuning on the domain dataset \benchmarksim{} leads to performance improvements on out-of-distribution tasks within the same domain, compared to the base model without fine-tuning. However, it remains uncertain whether fine-tuning on our domain dataset would also enhance performance on tasks from different domains, such as Python program synthesis tasks.

To investigate this, we evaluate our fine-tuned Llama3-8B-\sft{} model on other Python program synthesis benchmarks, including HumanEval~\citep{DBLP:journals/corr/abs-2107-03374}, HumanEval+~\citep{DBLP:conf/nips/LiuXW023}, MBPP~\citep{DBLP:journals/corr/abs-2108-07732}, and MBPP+~\citep{DBLP:conf/nips/LiuXW023}. Unlike our benchmarks, these benchmarks focus on general Python program synthesis tasks from natural language or docstrings, without visual elements present in the benchmark tasks.

\looseness-1
The results are presented in Figure~\ref{fig.ood_performance_program_synthesis_benchmarks}. Our findings indicate that fine-tuning on our domain dataset \benchmarksim{} results in a slight performance drop ($3\sim 6\%$) on these program synthesis benchmark tasks. We hypothesize that this is due to the \benchmarksim{} dataset's focus on visual programming tasks, which emphasize visual understanding, spatial reasoning, and planning—skills that are not directly applicable to other Python program synthesis tasks. Consequently, fine-tuning on our domain dataset does not provide additional knowledge for solving other benchmark tasks. Instead, the fine-tuning process may cause the model to forget some knowledge already acquired during the pre-training stage, leading to a performance drop in other benchmark tasks.

\subsection{Case Study: Output Code Analysis on Perturbed Tasks}

In the main paper, we presented a failure analysis by perturbing tasks and calculating the success rate. To illustrate the failure cases, we provide examples of output code from the evaluated models on these perturbed tasks, including GPT-4V, Llama3-70B, and Llama3-8B-\sft{}. 

The output code is displayed in Figure~\ref{fig.failure_examples}. In the provided examples, we observe that GPT-4V has difficulty handling grid constraints and spatial reasoning. For example, in \task{} and $\task{}_{\codecons{}}$, GPT-4V attempts to reach the strawberry by ignoring the walls. However, once the walls are removed ($\task{}_{\gridcons{}}$), GPT-4V is able to successfully solve the task. Interestingly, GPT-4V fails to solve $\task{}_{\codecons{}, \gridcons{}}$, even though this task is conceptually simpler than $\task{}_{\gridcons{}}$ due to the absence of code constraints. Upon examining the code and the accompanying comments from GPT-4V, we found that it miscalculates the strawberry's coordinates, indicating a struggle with spatial reasoning. Additionally, we observed that moving the turtle closer to the strawberry consistently improves GPT-4V's performance, suggesting that long-path planning and spatial reasoning are challenging for GPT-4V. However, for Llama3-70B and Llama3-8B-\sft{}, we observe that neither model successfully solves any of the perturbed tasks.


\begin{figure*}[t]
    \centering
    \adjustbox{scale=0.67}{
    \begin{tabular}{c c c c}
      \toprule
      Task & GPT-4V & Llama3-70B & Llama3-8B-\sft{} \\
      \midrule
      \begin{minipage}{0.17\textwidth}
        \centering
        \caption*{\scriptsize Find the strawberry. Use at most 8 commands.} \vspace{-8pt}
        \includegraphics[width=0.85\linewidth]{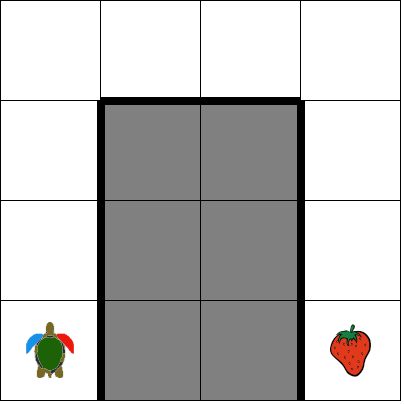}
        \caption*{\task{}}
      \end{minipage}
      &
      \begin{minipage}{0.45\textwidth}
      \begin{lstlisting}
def run():
  # Turtle faces east
  turn_right()

  # Moves to (1,3)
  $\textcolor{red}{\texttt{move\_forward()}}$

  # Moves to (2,3)
  move_forward() 

  # Moves to (3,3) where the strawberry is located
  move_forward() 
  $$
      \end{lstlisting}
      \end{minipage}
      &
      \begin{minipage}{0.22\textwidth}
        \begin{lstlisting}
def run():
  move_forward()
  turn_right()
  $\textcolor{red}{\texttt{move\_forward()}}$
  move_forward()
  turn_left()
  move_forward()
  move_forward()
  move_forward()
  $$
  $$
  $$
  $$
      \end{lstlisting}
      \end{minipage}
      &
      \begin{minipage}{0.22\textwidth}
        \begin{lstlisting}
def run():
  move_forward()
  move_forward()
  turn_right()
  for i in range(3):
    $\textcolor{red}{\texttt{move\_forward()}}$
  turn_right()
  for i in range(3):
    move_forward()
  $$
  $$
  $$
  $$
      \end{lstlisting}
      \end{minipage}
      \\
      \begin{minipage}{0.17\textwidth}
        \centering
        \caption*{\scriptsize \\ Find the strawberry.} \vspace{-8pt}
        \includegraphics[width=0.85\linewidth]{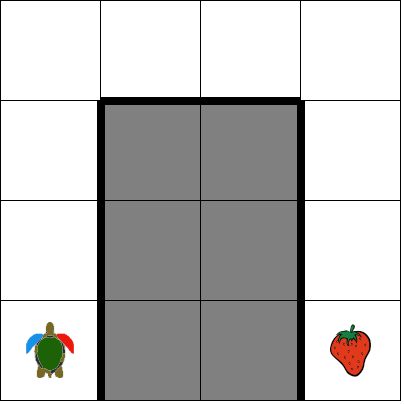}
        \caption*{$\task{}_{\codecons{}}$}
      \end{minipage}
      &
      \begin{minipage}{0.45\textwidth}
        \begin{lstlisting}
def run():
  # Start at (0,3) facing north
  # First, turn right to face east
  turn_right()
  
  # Move forward three times to reach (3,3)
  for i in range(3):
    $\textcolor{red}{\texttt{move\_forward()}}$
  $$
  $$
  $$
  $$
  $$
        \end{lstlisting}
      \end{minipage}
      &
      \begin{minipage}{0.22\textwidth}
        \begin{lstlisting}
def run():
  for i in range(3):
    $\textcolor{red}{\texttt{move\_forward()}}$
    turn_right()
  move_forward()
  turn_left()
  move_forward()
  turn_right()
  move_forward()
  turn_left()
  for i in range(2):
    move_forward()
  $$
        \end{lstlisting}
      \end{minipage}
      &
      \begin{minipage}{0.22\textwidth}
        \begin{lstlisting}
def run():
  move_forward()
  turn_right()
  $\textcolor{red}{\texttt{move\_forward()}}$
  turn_left()
  move_forward()
  move_forward()
  $$
  $$
  $$
  $$
  $$
  $$
        \end{lstlisting}
      \end{minipage}
      \\
      \begin{minipage}{0.17\textwidth}
        \centering
        \caption*{\scriptsize Find the strawberry. Use at most 8 commands.}
        \vspace{-8pt}
        \includegraphics[width=0.85\linewidth]{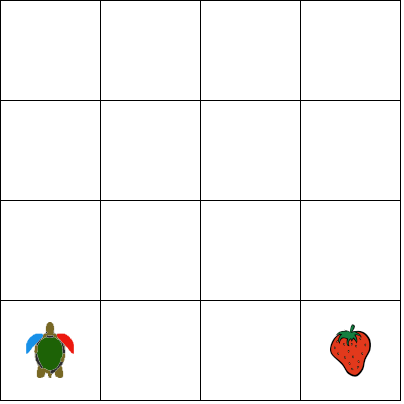}
        \caption*{$\task{}_{\gridcons}$}
      \end{minipage}
      &
      \begin{minipage}{0.45\textwidth}
        \begin{lstlisting}
def run():
  turn_right() # Turtle faces east
  for i in range(3):
    move_forward() # Moves from (0,3) to (3,3)
    $$
    $$
    $$
    $$
    $$
    $$
    $$
    $$
    $$
        \end{lstlisting}
      \end{minipage}
      &
      \begin{minipage}{0.22\textwidth}
        \begin{lstlisting}
def run():
  for i in range(3):
    $\textcolor{red}{\texttt{move\_backward()}}$
  turn_right()
  move_forward()
  turn_left()
  move_forward()
  move_forward()
  turn_right()
  move_forward()
  $$
  $$
  $$
        \end{lstlisting}
      \end{minipage}
      &
      \begin{minipage}{0.22\textwidth}
        \begin{lstlisting}
def run():
  move_forward()
  turn_right()
  for i in range(3):
    move_forward()
  turn_right()
  move_backward()
  for i in range(3):
    $\textcolor{red}{\texttt{move\_forward()}}$
  $$
  $$
  $$
  $$
        \end{lstlisting}
      \end{minipage}
      \\
      \begin{minipage}{0.17\textwidth}
        \centering
        \caption*{\scriptsize Find the strawberry. Use at most 8 commands.}
        \vspace{-8pt}
        \includegraphics[width=0.85\linewidth]{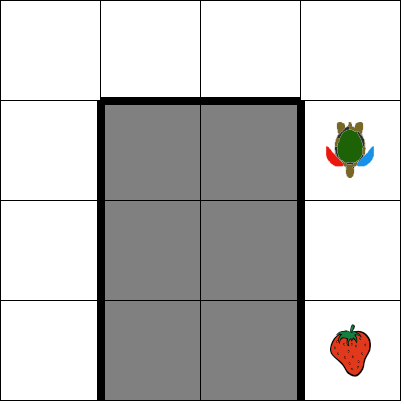}
        \caption*{$\task{}_{\spatial{}}$}
      \end{minipage}
      &
      \begin{minipage}{0.45\textwidth}
        \begin{lstlisting}
def run():
  move_forward() # Moves from (3,1) to (3,2)
  move_forward() # Moves from (3,2) to (3,3)
  $\hookrightarrow$ where the strawberry is located
  turn_right() # Turns to face west
  turn_right() # Turns to face north
  turn_right() # Turns to face east
  turn_right() # Turns to face south again
  turn_right() # Turns to face west
  turn_right() # Turns to face north    
  $$
  $$
  $$
        \end{lstlisting}
      \end{minipage}
      &
      \begin{minipage}{0.22\textwidth}
        \begin{lstlisting}
def run():
  turn_left()
  $\textcolor{red}{\texttt{move\_forward()}}$
  turn_right()
  move_forward()
  move_forward()
  turn_left()
  move_forward()
  move_forward()
  $$
  $$
  $$
  $$
        \end{lstlisting}
      \end{minipage}
      &
      \begin{minipage}{0.22\textwidth}
        \begin{lstlisting}
def run():
  move_forward()
  turn_right()
  for i in range(3):
    $\textcolor{red}{\texttt{move\_forward()}}$
  turn_right()
  move_forward()
  move_forward()
  $$
  $$
  $$
  $$
  $$
        \end{lstlisting}
      \end{minipage}
      \\
      \begin{minipage}{0.17\textwidth}
        \centering
        \caption*{\scriptsize \\Find the strawberry.} \vspace{-8pt}
        \includegraphics[width=0.85\linewidth]{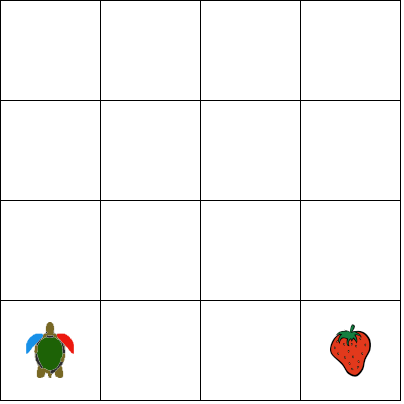}
        \caption*{$\task{}_{\codecons{}, \gridcons{}}$}
      \end{minipage}
      &
      \begin{minipage}{0.45\textwidth}
        \begin{lstlisting}
def run():
  # Turn right to face east
  turn_right()

  # Move forward three times to reach column 3
  for i in range(3):
    move_forward()

  # Turn right to face south
  turn_right()

  # Move forward to reach the strawberry at (3,3)
  $\textcolor{red}{\texttt{move\_forward()}}$
        \end{lstlisting}
      \end{minipage}
      &
      \begin{minipage}{0.22\textwidth}
        \begin{lstlisting}
def run():
  for i in range(3):
    move_forward()
  turn_right()
  move_forward()
  turn_left()
  for i in range(3):
    $\textcolor{red}{\texttt{move\_forward()}}$
  $$
  $$
  $$
  $$
  $$
      \end{lstlisting}
      \end{minipage}
      &
      \begin{minipage}{0.22\textwidth}
        \begin{lstlisting}
def run():
  move_forward()
  turn_right()
  move_forward()
  turn_right()
  move_forward()
  $\textcolor{red}{\texttt{move\_forward()}}$
  $$
  $$
  $$
  $$
  $$
  $$
        \end{lstlisting}
      \end{minipage}
      \\
      \begin{minipage}{0.17\textwidth}
        \centering
        \caption*{\scriptsize Find the strawberry. Use at most 8 commands.}
        \vspace{-8pt}
        \includegraphics[width=0.85\linewidth]{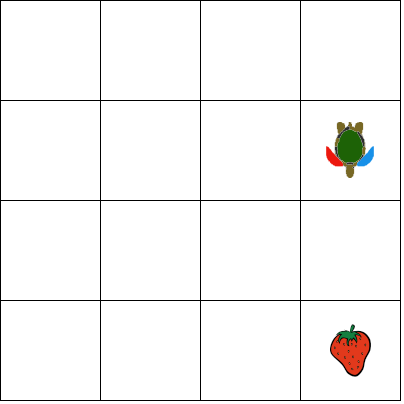}
        \caption*{$\task{}_{\gridcons{}, \spatial{}}$}
      \end{minipage}
      &
      \begin{minipage}{0.45\textwidth}
        \begin{lstlisting}
def run():
  # Moves from (3,1) to (3,2)
  move_forward()

  # Moves from (3,2) to (3,3)
  $\hookrightarrow$ where the strawberry is located
  move_forward() 
  $$
  $$
  $$
  $$
  $$
  $$
        \end{lstlisting}
      \end{minipage}
      &
      \begin{minipage}{0.22\textwidth}
        \begin{lstlisting}
def run():
  turn_left()
  turn_left()
  for i in range(2):
    $\textcolor{red}{\texttt{move\_forward()}}$
  turn_left()
  move_forward()
  turn_right()
  move_forward()
  $$
  $$
  $$
  $$
        \end{lstlisting}
      \end{minipage}
      &
      \begin{minipage}{0.22\textwidth}
        \begin{lstlisting}
def run():
  move_forward()
  turn_right()
  move_forward()
  for i in range(2):
    move_forward()
    turn_right()
  turn_left()
  $\textcolor{red}{\texttt{move\_forward()}}$
  $$
  $$
  $$
        \end{lstlisting}
      \end{minipage}
      \\
      \begin{minipage}{0.17\textwidth}
        \centering
        \caption*{\scriptsize \\ Find the strawberry.}
        \vspace{-8pt}
        \includegraphics[width=0.85\linewidth]{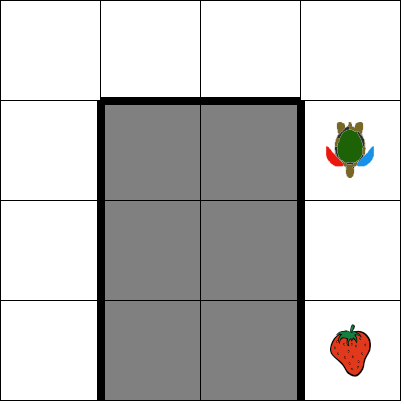}
        \caption*{$\task{}_{\codecons{}, \spatial{}}$}
      \end{minipage}
      &
      \begin{minipage}{0.45\textwidth}
        \begin{lstlisting}
def run():
  # The turtle starts at (3,1) facing south.
  # First, move forward to (3,2).
  move_forward()
  
  # Then, move forward to (3,3) 
  # where the strawberry is located.
  move_forward()
  $$
  $$
  $$
  $$
  $$
      \end{lstlisting}
      \end{minipage}
      &
      \begin{minipage}{0.22\textwidth}
        \begin{lstlisting}
def run():
  turn_left()
  $\textcolor{red}{\texttt{move\_forward()}}$
  turn_right()
  move_forward()
  turn_left()
  move_forward()
  turn_right()
  move_forward()
  turn_left()
  move_forward()
  turn_right()
  ... (repeat outputs)
        \end{lstlisting}
      \end{minipage}
      &
      \begin{minipage}{0.22\textwidth}
        \begin{lstlisting}
def run():
  move_forward()
  turn_right()
  $\textcolor{red}{\texttt{move\_forward()}}$
  turn_right()
  move_forward()
  move_forward()
  turn_left()
  move_forward()
  turn_right()
  move_forward()
  $$
  $$
        \end{lstlisting}
      \end{minipage}
      \\
      \begin{minipage}{0.17\textwidth}
        \centering
        \caption*{\scriptsize Find the strawberry.}
        \vspace{-8pt}
        \includegraphics[width=0.85\linewidth]{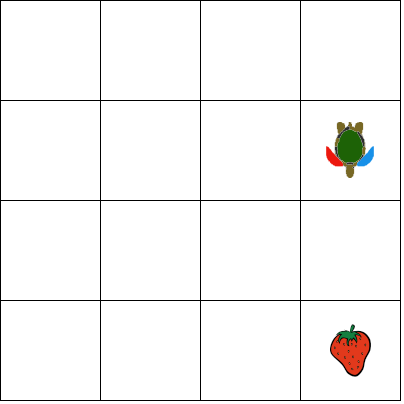}
        \caption*{$\task{}_{\codecons{}, \gridcons{}, \spatial{}}$}
      \end{minipage}
      &
      \begin{minipage}{0.45\textwidth}
        \begin{lstlisting}
def run():
  # The turtle starts at (3,1) facing south.
  # Move forward to reach (3,2).
  move_forward()
  
  # Move forward again to reach (3,3) 
  $\hookrightarrow$ where the strawberry is located.
  move_forward()
  $$
  $$
  $$
  $$
        \end{lstlisting}
      \end{minipage}
      &
      \begin{minipage}{0.22\textwidth}
        \begin{lstlisting}
def run():
  turn_left()
  turn_left()
  for i in range(2):
    $\textcolor{red}{\texttt{move\_forward()}}$
  turn_left()
  for i in range(3):
    move_forward()
  turn_right()
  move_forward()
  $$
  $$
        \end{lstlisting}
      \end{minipage}
      &
      \begin{minipage}{0.22\textwidth}
        \begin{lstlisting}
def run():
  turn_right()
  move_forward()
  turn_right()
  move_forward()
  turn_right()
  move_forward()
  $\textcolor{red}{\texttt{move\_forward()}}$
  turn_left()
  move_forward()
  move_forward()
  $$
        \end{lstlisting}
      \end{minipage}
      \\
      \bottomrule
    \end{tabular}
    }
    \caption{Output codes generated by GPT-4V, Llama3-70B, and Llama3-8B-\sft{} for various perturbations applied to a task $\task{}$. The perturbations include removing code constraints ($\task{}_{\codecons{}}$), removing grid constraints ($\task{}_{\gridcons{}}$), simplifying spatial relationships ($\task{}_{\spatial{}}$), and combinations of these perturbations ($\task{}_{\codecons{}, \gridcons{}}$, $\task{}_{\gridcons{}, \spatial{}}$, $\task{}_{\codecons{}, \spatial{}}$, and $\task{}_{\codecons{}, \gridcons{}, \spatial{}}$). Note that only the code is shown due to space limitations. The \textcolor{red}{red line} in the output code marks the point where the code first triggers an execution error or fails to successfully solve the task. GPT-4V successfully solves $5$ out of $8$ perturbed tasks, but Llama3-70B and fine-tuned Llama3-8B-\sft{} fail to solve any of the perturbed tasks.}
    \label{fig.failure_examples}
\end{figure*}

    \clearpage


\section{Prompts Used in the Benchmark} \label{sec:appendix:prompt}

In this section, we present three types of prompts for program synthesis in the \platformMini{} domain. Figures~\ref{fig.prompt_template_nl.example1} and~\ref{fig.prompt_template_ascii.example1} show examples of the prompts using natural language and ASCII representation, respectively. Figure~\ref{fig.prompt_template_nl.example_fewshot_cot} shows the prompt for the few-shot + CoT prompting.

\looseness-1Note that after the title ``\#\#\#\# Available Python Functions'' in prompts, we provide an explanation and two examples of the code format. This is intended for \emph{base models}, such as GPT-family and Llama-family base models, to ensure they follow the desired code format. However, \emph{fine-tuning models} does not need this code format in the prompt, as models are trained with formatted code directly. Therefore, we omit the code format and examples from the prompts when fine-tuning models.


\begin{figure*}[h!]
    \centering
    \setlength\tabcolsep{5pt}
    \renewcommand{\arraystretch}{1.2}
\tiny
\begin{tabular}{|p{1.0\linewidth}|}
    \hline
    \\[-6pt]
    \multicolumn{1}{|c|}{\promptheader{Natural Language Prompt for Code Generation in \platformMini{}}} \\
    \\[-6pt]
    You are presented with a visual programming task involving a goal, a grid, a turtle, various items (or lines). You need to write Python code that enables the turtle to accomplish the goal within the grid.
    \\\\
    \#\#\#\# Grid and Turtle
    \\
    - The task has a \`{}m x n\`{} grid. The coordinates of the grid cells are \`{}(x, y)\`{}, where \`{}x\`{} is the column number and \`{}y\`{} is the row number. The top-left cell has coordinates \`{}(0, 0)\`{}.
    - The turtle starts at a specific grid cell and faces one of four directions: North, East, South, or West.
    \\\\
    \#\#\#\# Items
    \\
    Each item in the grid is defined by three attributes:
    \\
    - \`{}count\`{}: The number of identical items in that grid cell.
    \\
    - \`{}color\`{}: The item's color. Options include red, green, blue, yellow, black, white, orange, purple, and pink.
    \\
    - \`{}name\`{}: The type of the item, such as circle, rectangle, triangle, cross, strawberry, or lemon.
    \\\\
    \#\#\#\# Lines
    \\
    Sometimes, the grid doesn't contain any items but has lines with colors. You need to draw lines of the specified color to solve the task.
    \\\\
    \#\#\#\# Grid Cell Properties\\
    - A grid cell may be \`{}accessible\`{} or \`{}forbidden\`{}. The turtle can move to an accessible cell but not into a forbidden cell. If the turtle tries to move into a forbidden cell, it will crash and fail to solve the task.\\
    - Grid cells can have walls on their edges (top, bottom, left, and right). The turtle cannot move through walls, otherwise it will crash and fail to solve the task.
    \\\\
    \#\#\#\# Available Python Functions\\
    To solve the task, you can use the following Python functions:\\
    - \`{}move\_forward()\`{}: This function moves the turtle forward in the direction it is facing by one grid cell. For example, if the turtle is at the position (x, y) and facing north, after executing move\_forward(), the turtle will be at the position (x, y-1).  \\
    - \`{}move\_backward()\`{}: This function moves the turtle backward in the direction it is facing by one grid cell. For example, if the turtle is at the position (x, y) and facing west, after executing \`{}move\_backward()\`{}, the turtle will be at the position (x+1, y).  \\
    - \`{}turn\_left()\`{}: This function makes the turtle turn left in the direction it is facing - by 90 degrees. For example, if the turtle is facing north, after executing \`{}turn\_left()\`{}, the turtle will be facing west.  \\
    - \`{}turn\_right()\`{}: This function makes the turtle turn right in the direction it is facing - by 90 degrees. For example, if the turtle is facing south, after executing \`{}turn\_right()\`{}, the turtle will be facing west.  \\
    - \`{}setpc(color)\`{}: This function sets the pen color to the specified color. The available colors are: red, green, blue, yellow, black, white. The default pen color is black. The trajectory of the turtle is drawn with the pen color.  \\
    - \`{}for\`{} loop: This loop is used to repeat a set of commands a specified number of times. For example, \`{}for i in range(4):\`{} will repeat the commands inside the loop 4 times.  \\
    Your code should follow the format:\\
    \`{}\`{}\`{}python\\
    def run():\\
    \quad \# Your solution code goes here\\
    \quad pass\\
    \`{}\`{}\`{}\\
    Here are some examples of the code:\\
    Example 1:\\
    \`{}\`{}\`{}python\\
    def run():\\
        \quad move\_forward()\\
        \quad for i in range(4):\\
            \quad \quad move\_forward()\\
            \quad \quad turn\_left()\\
    \`{}\`{}\`{}
    \\
    Example 2:\\
    \`{}\`{}\`{}python\\
    def run():\\
        \quad move\_forward()\\
        \quad setpc('red')\\
        \quad for i in range(3):\\
            \quad \quad move\_forward()\\
        \quad turn\_right()\\
        \quad move\_backward()\\
    \`{}\`{}\`{}\\
    \\
    Now, write a CORRECT Python code that successfully solves the following task.
    \\
    \#\#\# Task:\\
    A 3x3 grid. The turtle starts at (1,1) facing north.
    \\
    Accessible cells: (0,0), (1,0), (2,0), (0,1), (1,1), (2,1), (0,2), (1,2), (2,2).
    \\
    Items in the grid:\\
    - 1 red strawberry at (1,0).
    \\\\
    \#\#\# Goal:\\
    Find the strawberry.
    \\\\
    \#\#\# CORRECT code:
    \\
    \hline
\end{tabular}

    \caption{An example of natural language prompt in the \platformMini{} domain.}
    \label{fig.prompt_template_nl.example1}
\end{figure*}


\begin{figure*}[h!]
    \centering
    \setlength\tabcolsep{5pt}
    \renewcommand{\arraystretch}{1.2}
\tiny
\begin{tabular}{|p{1\linewidth}|}
    \hline
    \\[-6pt]
    \multicolumn{1}{|c|}{\promptheader{ASCII-based Prompt for Program Synthesis in \platformMini{}}} \\
    \\[-6pt]
    You are presented with a visual programming task involving a goal, a grid, a turtle, various items (or lines). You need to write Python code that enables the turtle to accomplish the goal within the grid. \\\\

    \#\#\#\# Grid and Turtle\\
    A task's grid contain a turtle and some items. The turtle can face one of four directions: North (\`{}\texttt{\^}\`{}), South (\`{}v\`{}), East (\`{}>\`{}), or West (\`{}<\`{}). An item has three attributes: \`{}count\`{}, \`{}color\`{}, and \`{}name\`{}. The \`{}count\`{} indicates the number of identical items in that grid cell. The \`{}color\`{} specifies the item's color, and the \`{}name\`{} describes the item's type. Here are the possible options:\\
    - Colors: Red (\`{}R\`{}), Green (\`{}G\`{}), Blue (\`{}B\`{}), Yellow (\`{}Y\`{}), Black (\`{}K\`{}), White (\`{}W\`{}), Orange (\`{}O\`{}), Purple (\`{}U\`{}), Pink (\`{}P\`{})\\
    - Names: Circle (\`{}o\`{}), Rectangle (\`{}$\square$\`{}), Triangle (\`{}$\triangle$\`{}) ,Cross (\`{}X\`{}), Strawberry (\`{}S\`{}), Lemon (\`{}L\`{}) \\
    - Counts: \`{}1\`{}, \`{}2\`{}, \`{}3\`{}, \`{}4\`{}\\
    - For example, \`{}2RS\`{} means two red strawberries.
    \\\\
    We use the following symbols to describe a grid:\\
    - \`{}---\`{} represents the top or bottom edge of a grid cell.\\
    - \`{}|\`{} represents the left or right edge of a grid cell.\\
    - \`{}===\`{} represents an upper or lower wall of a cell. \\
    - \`{}‖\`{} represents a left or right wall of a cell.\\
    - \`{}+\`{} represents the corner of a grid cell.\\
    - \`{}X\`{} represents a forbidden cell that cannot be accessed.\\
    \\
    \#\#\#\# Grid Cell Properties\\
    - A grid cell may be \`{}accessible\`{} or \`{}forbidden\`{}. The turtle can move to an accessible cell but not into a forbidden cell. If the turtle tries to move into a forbidden cell, it will crash and fail to solve the task.\\
    - Grid cells can have walls on their edges (top, bottom, left, and right). The turtle cannot move through walls, otherwise it will crash and fail to solve the task.
    \\\\
    \#\#\#\# Available Python Functions\\

    To solve the task, you can use the following Python functions:\\
    - \`{}move\_forward()\`{}: This function moves the turtle forward in the direction it is facing by one grid cell. For example, if the turtle is at the position (x, y) and facing north, after executing move\_forward(), the turtle will be at the position (x, y-1).  \\
    - \`{}move\_backward()\`{}: This function moves the turtle backward in the direction it is facing by one grid cell. For example, if the turtle is at the position (x, y) and facing west, after executing \`{}move\_backward()\`{}, the turtle will be at the position (x+1, y).  \\
    - \`{}turn\_left()\`{}: This function makes the turtle turn left in the direction it is facing - by 90 degrees. For example, if the turtle is facing north, after executing \`{}turn\_left()\`{}, the turtle will be facing west.  \\
    - \`{}turn\_right()\`{}: This function makes the turtle turn right in the direction it is facing - by 90 degrees. For example, if the turtle is facing south, after executing \`{}turn\_right()\`{}, the turtle will be facing west.  \\
    - \`{}setpc(color)\`{}: This function sets the pen color to the specified color. The available colors are: red, green, blue, yellow, black, white. The default pen color is black. The trajectory of the turtle is drawn with the pen color.  \\
    - \`{}for\`{} loop: This loop is used to repeat a set of commands a specified number of times. For example, \`{}for i in range(4):\`{} will repeat the commands inside the loop 4 times.  \\
    Your code should follow the format:\\
    \`{}\`{}\`{}python\\
    def run():\\
    \quad \# Your solution code goes here\\
    \quad pass\\
    \`{}\`{}\`{}\\
    Here are some examples of the code:\\
    Example 1:\\
    \`{}\`{}\`{}python\\
    def run():\\
        \quad move\_forward()\\
        \quad for i in range(4):\\
            \quad \quad move\_forward()\\
            \quad \quad turn\_left()\\
    \`{}\`{}\`{}
    \\
    Example 2:\\
    \`{}\`{}\`{}python\\
    def run():\\
        \quad move\_forward()\\
        \quad setpc('red')\\
        \quad for i in range(3):\\
            \quad \quad move\_forward()\\
        \quad turn\_right()\\
        \quad move\_backward()\\
    \`{}\`{}\`{}\\
    \\
    Now, write a CORRECT Python code that successfully solves the following task:\\
    \#\#\# Task:\begin{verbatim}
+---+---+---+
|   |1RS|   |
+---+---+---+
|   | ^ |   |
+---+---+---+
|   |   |   |
+---+---+---+\end{verbatim}\#\#\# Goal:\\
    Find the strawberry.\\

    \#\#\# CORRECT Code:\\
    \hline
\end{tabular}

    \caption{An example of ASCII-based prompt in the \platformMini{} domain.}
    \label{fig.prompt_template_ascii.example1}
\end{figure*}


\begin{figure*}[h!]
    \centering
    \setlength\tabcolsep{5pt}
    \renewcommand{\arraystretch}{1.2}
\tiny
\begin{tabular}{|p{1.0\linewidth}|}
    \hline
    \\[-6pt]
    \multicolumn{1}{|c|}{\promptheader{Few-shot + CoT Prompt for Code Generation in \platformMini{}}} \\
    \\[-6pt]
    You are presented with a visual programming task involving a goal, a grid, a turtle, various items (or lines). You need to write Python code that enables the turtle to accomplish the goal within the grid.
    \\\\
    \#\#\#\# Grid and Turtle
    \\
    - The task has a \`{}m x n\`{} grid. The coordinates of the grid cells are \`{}(x, y)\`{}, where \`{}x\`{} is the column number and \`{}y\`{} is the row number. The top-left cell has coordinates \`{}(0, 0)\`{}.
    - The turtle starts at a specific grid cell and faces one of four directions: North, East, South, or West.
    \\\\
    \#\#\#\# Items
    \\
    Each item in the grid is defined by three attributes:
    \\
    - \`{}count\`{}: The number of identical items in that grid cell.
    \\
    - \`{}color\`{}: The item's color. Options include red, green, blue, yellow, black, white, orange, purple, and pink.
    \\
    - \`{}name\`{}: The type of the item, such as circle, rectangle, triangle, cross, strawberry, or lemon.
    \\\\
    \#\#\#\# Lines
    \\
    Sometimes, the grid doesn't contain any items but has lines with colors. You need to draw lines of the specified color to solve the task.
    \\\\
    \#\#\#\# Grid Cell Properties\\
    - A grid cell may be \`{}accessible\`{} or \`{}forbidden\`{}. The turtle can move to an accessible cell but not into a forbidden cell. If the turtle tries to move into a forbidden cell, it will crash and fail to solve the task.\\
    - Grid cells can have walls on their edges (top, bottom, left, and right). The turtle cannot move through walls, otherwise it will crash and fail to solve the task.
    \\\\
    \#\#\#\# Available Python Functions\\
    To solve the task, you can use the following Python functions:\\
    - \`{}move\_forward()\`{}: This function moves the turtle forward in the direction it is facing by one grid cell. For example, if the turtle is at the position (x, y) and facing north, after executing move\_forward(), the turtle will be at the position (x, y-1).  \\
    - \`{}move\_backward()\`{}: This function moves the turtle backward in the direction it is facing by one grid cell. For example, if the turtle is at the position (x, y) and facing west, after executing \`{}move\_backward()\`{}, the turtle will be at the position (x+1, y).  \\
    - \`{}turn\_left()\`{}: This function makes the turtle turn left in the direction it is facing - by 90 degrees. For example, if the turtle is facing north, after executing \`{}turn\_left()\`{}, the turtle will be facing west.  \\
    - \`{}turn\_right()\`{}: This function makes the turtle turn right in the direction it is facing - by 90 degrees. For example, if the turtle is facing south, after executing \`{}turn\_right()\`{}, the turtle will be facing west.  \\
    - \`{}setpc(color)\`{}: This function sets the pen color to the specified color. The available colors are: red, green, blue, yellow, black, white. The default pen color is black. The trajectory of the turtle is drawn with the pen color.  \\
    - \`{}for\`{} loop: This loop is used to repeat a set of commands a specified number of times. For example, \`{}for i in range(4):\`{} will repeat the commands inside the loop 4 times.  \\
    \\
    Your code should follow the format:\\
    \`{}\`{}\`{}python\\
    def run():\\
    \quad \# Your solution code goes here\\
    \quad pass\\
    \`{}\`{}\`{}\\\\

    Here are some examples of the the tasks and their corresponding solution codes: \\

    \textcolor{blue}{\{few\_shot\_example\_1\}} \\
    \textcolor{blue}{\{few\_shot\_example\_2\}} \\
    \textcolor{blue}{\{few\_shot\_example\_3\}} \\

    \\
    Now, write a CORRECT Python code that successfully solves the following task.
    \\
    \#\#\# Task:\\
    A 3x3 grid. The turtle starts at (1,1) facing north.
    \\
    Accessible cells: (0,0), (1,0), (2,0), (0,1), (1,1), (2,1), (0,2), (1,2), (2,2).
    \\
    Items in the grid:\\
    - 1 red strawberry at (1,0).
    \\\\
    \#\#\# Goal:\\
    Find the strawberry.
    \\\\
    \#\#\# CORRECT code:
    \\
    \hline
\end{tabular}

    \caption{An example of the few-shot + CoT prompt in the \platformMini{} domain. This prompt has several \promptinput{placeholders} to include details for few-shot examples.}
    \label{fig.prompt_template_nl.example_fewshot_cot}
\end{figure*}

\begin{figure*}[h!]
    \centering
    \setlength\tabcolsep{5pt}
    \renewcommand{\arraystretch}{1.2}
\tiny
\begin{tabular}{|p{1.0\linewidth}|}
    \hline
    \\[-6pt]
    \multicolumn{1}{|c|}{\promptheader{\{few\_shot\_example\_1\}}} \\
    \\[-6pt]
    \#\#\# Task 1:
    A 4x4 grid. The turtle starts at (3,3) facing south.\\

    Accessible cells: (3,0), (3,1), (3,2), (0,3), (1,3), (2,3), (3,3).\\
    Forbidden cells: (0,0), (1,0), (2,0), (0,1), (1,1), (2,1), (0,2), (1,2), (2,2).\\

    Lines in the grid:\\
    - A red line from (3,0) to (3,1).\\
    - A red line from (3,1) to (3,2).\\
    - A red line from (3,2) to (3,3).\\
    - A red line from (0,3) to (1,3).\\
    - A red line from (1,3) to (2,3).\\
    - A red line from (2,3) to (3,3).\\
    \\
    \#\#\# Goal 1: \\
    Draw the picture in red. Use at most 8 commands.\\
    \\
    \#\#\# Code 1:\\
    Explanation:\\
    1. Starting Position and Initial Setup:\\
        - The turtle starts at \`{}(3, 3)\`{}, facing south. To achieve the goal of drawing the picture in red, we first need to set the pen color to red using the \`{}setpc('red')\`{} command.\\\\

    2. Plan the path:\\
        - To draw the pattern, the path is \`{}(3,3) -> (3,0) -> (3,3) -> (0,3)\`{}.\\
    \\
    3. Compiling the Path into Commands:\\
        - The turtle starts at \`{}(3, 3)\`{}, facing south. To draw the first red line from \`{}(3, 3)\`{} to \`{}(3, 0)\`{}, we need to first set the pen color to red and then move back 3 times to reach from \`{}(3, 3)\`{} to \`{}(3, 0)\`{}.\\
        - Now, the turtle is at \`{}(3, 0)\`{} and still facing south. To move back to \`{}(3, 3)\`{}, we move forward 3 times in the same direction (south).\\
        - At \`{}(3, 3)\`{} and facing south, the turtle needs to turn right to face west.\\
        - Now, the turtle is at \`{}(3, 3)\`{} and facing west. Move forward 3 times to reach from \`{}(3, 3)\`{} to \`{}(0, 3)\`{} facing west.\\
        - Now the turtle has drawn the picture in red.\\
    \\
    Putting it all together and notice that the solution code can use at most 8 commands. Here is the solution code:\\
    \\
    \`{}\`{}\`{}python\\
    def run():\\
    \quad setpc('red')\\
    \quad for i in range(3):\\
    \quad \quad move\_backward()\\
    \quad for i in range(3):\\
    \quad \quad move\_forward()\\
    \quad turn\_right()\\
    \quad for i in range(3):\\
    \quad \quad move\_forward()\\
    \`{}\`{}\`{}\\
    \hline
\end{tabular}

    \caption{The detail of the first few-shot + CoT example \textcolor{blue}{\{few\_shot\_example\_1\}}.}
    \label{fig.prompt_template_nl.example1_fewshot_cot}
\end{figure*}

\begin{figure*}[h!]
    \centering
    \setlength\tabcolsep{5pt}
    \renewcommand{\arraystretch}{1.2}
\tiny
\begin{tabular}{|p{1.0\linewidth}|}
    \hline
    \\[-6pt]
    \multicolumn{1}{|c|}{\promptheader{\{few\_shot\_example\_2\}}} \\
    \\[-6pt]
    \#\#\# Task 2:\\
    A 3x3 grid. The turtle starts at (1,2) facing south.\\\\

    Accessible cells: (0,0), (1,0), (2,0), (0,1), (1,1), (2,1), (0,2), (1,2), (2,2).\\
    (1,0) has a wall at the bottom edge.\\
    (2,0) has a wall at the bottom edge.\\
    (1,1) has a wall at the top edge.\\
    (2,1) has a wall at the top edge.\\
    \\
    Items in the grid:\\
    - 4 red strawberries at (1,0).\\
    - 1 red strawberry at (2,0).\\
    - 2 red strawberries at (2,1).\\
    - 1 red strawberry at (2,2).\\
    \\
    \#\#\# Goal 2: \\
    Collect exactly 5 strawberries.\\
    \\
    \#\#\# Code 2:\\
    Explanation:\\
    1. Starting Position and Initial Setup:\\
    - The turtle starts at \`{}(1, 2)\`{}, facing south.\\
    - The goal is to collect exactly 5 strawberries.\\
    - There are 4 strawberries at \`{}(1, 0)\`{} and 1 strawberry at \`{}(2, 0)\`{}. By visiting these two cells, we can collect exactly 5 strawberries. We need a path to visit both \`{}(1, 0)\`{} and \`{}(2, 0)\`{} while considering walls that block movement between cells \`{}(1, 0)\`{} \& \`{}(1, 1)\`{} and \`{}(2, 0)\`{} \& \`{}(2, 1)\`{}.\\
    \\
    2. Planning the Path:\\
    - An ideal path to avoid the walls and collect strawberries is: \`{}(1, 2) -> (1, 1) -> (0, 1) -> (0, 0) -> (2, 0)\`{}.\\
    - This path efficiently collects exactly 5 strawberries.\\
    \\
    3. Compiling the Path into Commands:\\
    - The turtle starts at \`{}(1, 2)\`{} facing south.\\
    - To move from \`{}(1, 2)\`{} to \`{}(1, 1)\`{}, it needs to move backward. Now the turtle is at \`{}(1, 1)\`{} facing south.\\
    - To move from \`{}(1, 1)\`{} to \`{}(0, 1)\`{}, it needs to turn right and move forward. Now it is at \`{}(0, 1)\`{} facing west.\\
    - To move from \`{}(0, 1)\`{} to \`{}(0, 0)\`{}, it needs to turn right to face north, then move forward. Now it is at \`{}(0, 0)\`{} facing north.\\
    - To move from \`{}(0, 0)\`{} to \`{}(2, 0)\`{}, it needs to turn right to face east, and finally, move forward twice to go from \`{}(0, 0)\`{} to \`{}(2, 0)\`{}.\\
    - Now the turtle has collected exactly 5 strawberries.\\
    \\
    Putting it all together. The solution code is:\\ \\

    \`{}\`{}\`{}python\\
    def run():\\
    \quad move\_backward()\\
    \quad turn\_right()\\
    \quad move\_forward()\\
    \quad turn\_right()\\
    \quad move\_forward()\\
    \quad turn\_right()\\
    \quad move\_forward()\\
    \quad move\_forward()\\
    \`{}\`{}\`{}\\
    \hline
\end{tabular}

    \caption{The detail of the second few-shot + CoT example \textcolor{blue}{\{few\_shot\_example\_2\}}.}
    \label{fig.prompt_template_nl.example2_fewshot_cot}
\end{figure*}

\begin{figure*}[h!]
    \centering
    \setlength\tabcolsep{5pt}
    \renewcommand{\arraystretch}{1.2}
\tiny
\begin{tabular}{|p{1.0\linewidth}|}
    \hline
    \\[-6pt]
    \multicolumn{1}{|c|}{\promptheader{\{few\_shot\_example\_3\}}} \\
    \\[-6pt]
    \#\#\# Task 3:\\
    A 3x3 grid. The turtle starts at (2,2) facing south.\\

    Accessible cells: (0,0), (1,0), (2,0), (0,1), (1,1), (2,1), (0,2), (1,2), (2,2).\\

    Items in the grid:\\
    - 1 blue triangle at (0,0).\\
    - 1 red circle at (1,0).\\
    - 1 green circle at (2,0).\\
    - 1 blue circle at (0,1).\\
    - 1 blue rectangle at (1,1).\\
    - 1 red triangle at (2,1).\\
    - 1 red cross at (0,2).\\
    - 1 green rectangle at (1,2).\\\\

    \#\#\# Goal 3: \\
    Find the blue triangle without standing on a red. Use at most 3 'backward'.\\\\

    \#\#\# Code 3:\\
    Explanation:\\
    1. Starting Position and Initial Setup:\\
    - The turtle starts at \`{}(2, 2)\`{}, facing south.\\
    - The goal is to find the blue triangle located at \`{}(0, 0)\`{} without standing on any red items.\\
    - There are several items in the grid: red circle at \`{}(1, 0)\`{}, red triangle at \`{}(2, 1)\`{}, and red cross at \`{}(0, 2)\`{}.\\
    - We need to avoid these red items and reach \`{}(0, 0)\`{} using at most 3 backward movements.\\\\
    2. Planning the Path:\\
    - Since we have to avoid red items and should use at most 3 backward movements, we need to plan an efficient path.\\
    - A suitable path that avoids red items and stays within the permitted movements is: \`{}(2, 2) -> (1, 2) -> (1, 1) -> (0, 1) -> (0, 0)\`{}.\\\\
    3. Compiling the Path into Commands:\\
    - The turtle starts at \`{}(2, 2)\`{} facing south. First, turn left to face east.\\
    - Move backward to reach \`{}(1, 2)\`{} facing east.\\
    - Turn left again to face north.\\
    - Move forward to reach \`{}(1, 1)\`{} facing north.\\
    - Turn right to face east.\\
    - Move backward to reach \`{}(0, 1)\`{} facing east.\\
    - Turn right to face south.\\
    - Move backward to reach \`{}(0, 0)\`{} facing south.\\
\\
    Putting it all together. The solution code is:\\
\\
    \`{}\`{}\`{}python\\
    def run():\\
            turn\_left()\\
            move\_backward()\\
            turn\_left()\\
            move\_forward()\\
            turn\_right()\\
            move\_backward()\\
            turn\_right()\\
            move\_backward()\\
    \`{}\`{}\`{}\\
    \hline
\end{tabular}

    \caption{The detail of the third few-shot + CoT example \textcolor{blue}{\{few\_shot\_example\_3\}}.}
    \label{fig.prompt_template_nl.example3_fewshot_cot}
\end{figure*}
}

}
{
}

\iftoggle{MainContentOnly}{

\begin{abstract}
Large language and multimodal models have shown remarkable success on various benchmarks focused on specific skills such as general-purpose programming, math word problem-solving, and visual question answering. However, it is unclear how well these models perform on tasks that require a combination of these skills. In this paper, we curate a novel program synthesis benchmark based on the real-world tasks in the XLogoOnline visual programming environment. Each task requires a combination of different skills such as spatial planning, basic programming, and logical reasoning. Our evaluation shows that current state-of-the-art models like GPT-4V and Llama3-70B struggle to solve these tasks, achieving only $20\%$ and $2.35\%$ success rates, respectively. Next, we develop a fine-tuning pipeline to boost the performance of models by leveraging a large-scale synthetic training dataset with over $80,000$ tasks. Moreover, we showcase how emulator-driven feedback can be used to design a curriculum over training data distribution, through which a fine-tuned Llama3-8B drastically outperforms GPT-4V and Llama3-70B models. Finally, we provide an in-depth failure analysis to understand the limitations of different models. We will publicly release the benchmark for future research on program synthesis in visual programming. 
\end{abstract}



\section{Concluding Discussions}
\label{sec.conclusion}

In this paper, we introduced \benchmark{}, a visual programming benchmark designed to evaluate the program synthesis capabilities of large models on visual programming tasks using the \platformAll{} environment. We found that large models struggle with visual programming tasks that require a combination of skills, even though our benchmark tasks require only basic programming skills. To improve performance, we developed a fine-tuning pipeline that involves synthetic dataset generation followed by supervised fine-tuning. This pipeline enabled the Llama3-8B model to achieve a success rate of $54.12\%$ on the benchmark tasks. Additionally, we demonstrated that leveraging emulator-driven feedback can further enhance standard fine-tuning performance by approximately $6\%$ in both the Llama3-8B and Llama2-7B models. 
Our failure analysis revealed that different models struggle with distinct issues: DeepSeek-R1-Distill-Llama-70B struggles with recursive reasoning, GPT-4V and Llama3-70B with spatial reasoning, and the fine-tuned Llama3-8B-\sft{} with grid constraints.



\section*{Limitations}

We discuss some limitations of our work and propose ideas for addressing them in the future. 
First, our benchmark focuses on basic programming skills, and future work could extend it to include more complex programming tasks. This could involve tasks that require more advanced programming concepts, such as conditionals and functions. 
Second, our emulator-driven fine-tuning provides the model with only binary feedback on the correctness of the predicted code. In the future, it would be interesting to provide more detailed feedback, such as identifying specific errors in the generated code and then using this more informative feedback to guide the fine-tuning process. 

\section*{Acknowledgments}
Funded/Co-funded by the European Union (ERC, TOPS, 101039090). Views and opinions expressed are however those of the author(s) only and do not necessarily reflect those of the European Union or the European Research Council. Neither the European Union nor the granting authority can be held responsible for them.
    \bibliography{main}
}
{
}

\iftoggle{SuppContentOnly}{
    \setcounter{figure}{9}
    
    \bibliography{main}    
}
{
}

\end{document}